\documentclass[11pt]{article}

\usepackage[margin=1in]{geometry}
\usepackage{amsmath, amssymb, amsthm}
\usepackage{mathtools}
\usepackage{booktabs}
\usepackage{array}
\usepackage{tabularx}
\usepackage{xcolor}
\usepackage{hyperref}
\usepackage{microtype}
\usepackage{parskip}
\usepackage{enumitem}
\usepackage{fancyhdr}
\usepackage{titlesec}
\usepackage{algorithm}
\usepackage{algpseudocode}
\usepackage{listings}
\usepackage{caption}
\usepackage{graphicx}
\usepackage{mdframed}
\usepackage{tcolorbox}
\usepackage{natbib}
\usepackage{tikz}
\usetikzlibrary{positioning, fit, backgrounds, arrows.meta, calc, decorations.pathreplacing, shapes.geometric}
\usepackage[section]{placeins}
\usepackage{longtable}

\newcommand{\simtype}[1]{{\tiny\ttfamily\color{gray!55}#1}}
\hypersetup{
	colorlinks=true,
	linkcolor=blue!60!black,
	citecolor=green!50!black,
	urlcolor=blue!60!black,
}

\definecolor{codebg}{RGB}{245, 245, 245}
\definecolor{codefg}{RGB}{40, 40, 40}

\DeclareMathOperator*{\argmin}{arg\,min}
\newcommand{\indicator}[1]{\mathbb{I}\!\left[#1\right]}
\newcommand{\Neigh}{\mathcal{N}}
\newcommand{\KP}{\mathcal{K}}

\titleformat{\section}{\large\bfseries}{\thesection.}{0.5em}{}
\titleformat{\subsection}{\normalsize\bfseries}{\thesubsection}{0.5em}{}
\titleformat{\subsubsection}{\normalsize\bfseries\itshape}{\thesubsubsection}{0.5em}{}

\title{%
	\vspace{-1em}
	\textbf{OrgForge: A Multi-Agent Simulation Framework\\
		for Verifiable Synthetic Organizational Corpora}\\[0.4em]
	\large\normalfont A General Architecture for Ground-Truth-Guaranteed\\
	Synthetic Data Across Enterprise AI Evaluation Domains
}

\author{
	Jeffrey Flynt\\
	\textit{Independent Researcher}\\[0.3em]
	\href{mailto:jeffrey.flynt@utexas.edu}{\texttt{jeffrey.flynt@utexas.edu}}\\[0.2em]
}

\date{March 2026}

\begin{document}
	\maketitle
	\thispagestyle{fancy}

\begin{abstract}
	Building and evaluating enterprise AI systems requires synthetic organizational corpora that are internally consistent, temporally structured, and cross-artifact traceable. Existing corpora either carry legal constraints or inherit hallucination artifacts from the generating LLMs, silently corrupting results when timestamps or facts contradict across documents, and reinforcing those errors during training.
	
	We present \textsc{OrgForge}, an open-source multi-agent simulation framework that enforces a strict physics-cognition boundary: a deterministic Python engine maintains a SimEvent ground-truth bus while LLMs generate only surface prose. \textsc{OrgForge} simulates the organizational processes that produce documents, not the documents themselves. Engineers leave mid-sprint, triggering incident handoffs and CRM ownership lapses. Knowledge gaps emerge when under-documented systems break and recover through organic documentation and incident resolution. Customer emails fire only when simulation state warrants contact; silence is verifiable ground truth. A live CRM state machine extends the physics-cognition boundary to the customer boundary, producing cross-system causal cascades spanning engineering incidents, support escalation, deal risk flagging, and SLA-adjusted invoices.
	
	The framework generates fifteen interleaved artifact categories traceable to a shared immutable event log. Four graph-dynamic subsystems govern organizational behavior independently of any LLM. An embedding-based ticket assignment system using the Hungarian algorithm makes the simulation domain-agnostic. An empirical evaluation across ten incidents demonstrates a 0.46 absolute improvement in prose-to-ground-truth fidelity over chained LLM baselines, and isolates a consistent hallucination failure mode in which chaining propagates fabricated facts faithfully across documents without correcting them. \textsc{OrgForge} is available under the MIT license.
\end{abstract}
	
	\noindent\rule{\linewidth}{0.4pt}
	
	\noindent\textbf{Availability.}
	Code and corpus generation tools: \href{https://github.com/aeriesec/orgforge}{\texttt{github.com/aeriesec/orgforge}}.
	Dataset: \href{https://huggingface.co/datasets/aeriesec/orgforge}{\texttt{huggingface.co/datasets/aeriesec/orgforge}}.
	Archived release: \href{https://zenodo.org/records/19036018}{\texttt{zenodo.org/records/19036018}}.
	
	\noindent\rule{\linewidth}{0.4pt}

	\section{Introduction}
	
	Enterprise AI systems, ranging from retrieval-augmented generation (RAG)
	pipelines to compliance tools, organizational agents, and agentic development
	sandboxes, share a fundamental data challenge: they require corpora with
	knowable, temporally structured, and cross-artifact traceable ground truth for
	development, training, and evaluation alike. Real-world datasets rarely provide
	these three properties simultaneously, and when they do, they carry legal
	constraints that limit redistribution and use. Furthermore, purely synthetic
	data generated by Large Language Models (LLMs) introduces a subtle failure
	mode where the generating model's hallucinations result in factual
	contradictions across documents, a problem that corrupts evaluation benchmarks
	and quietly degrades fine-tuned models.
	
\subsection{Limitations of Current Benchmarks}
\label{subsec:limitations-benchmarks}

The RAG evaluation problem illustrates this gap sharply. Current benchmarks
typically evaluate retrieval quality in isolation or test end-to-end question
answering on static passages. Neither approach captures the nuances of real
enterprise knowledge bases, which are characterized by documents that reference
each other across systems, facts that evolve over time, and incidents that
leave traces across multiple artifact types.

\subsection{Limitations of Synthetic Training Corpora}
\label{subsec:limitations-training}

The training data problem is structurally identical. Self-Instruct and related
approaches demonstrate that LLM-generated synthetic data can match curated
human data for training, but only when factual consistency across examples can
be assumed. In organizational corpora, where the same incident should appear in
a Slack thread, a JIRA ticket, a postmortem, and an invoice, that assumption
does not hold for LLM-generated data without an external consistency enforcer.
The same cross-document inconsistency that corrupts benchmarks silently degrades
fine-tuned models: a hallucinated fact is noise in a single example, but noise
that training can reinforce.

\subsection{Adjacent Use Cases}
\label{subsec:limitations-adjacent}

The same data deficit affects systems that are neither purely evaluative nor
purely generative. Teams building organizational agents need a live environment
with verifiable ground truth to score decisions against before deployment.
Security and compliance tooling (DLP systems, SIEM pipelines, insider threat
detection) requires realistic cross-system organizational data with known
labels, which real corpora cannot provide without legal constraint and synthetic
corpora cannot provide without a consistency enforcer. Organizational behavior
researchers studying stress propagation, knowledge degradation, and
communication pattern evolution lack a simulation substrate that produces
verifiable documentary artifacts alongside behavioral outputs.
	
\subsection{The OrgForge Framework}
\label{subsec:orgforge}

OrgForge addresses these limitations through a strict architectural boundary
where LLMs propose while the engine executes: a deterministic simulation engine
controls all underlying facts (on-call rotations, incident timelines, ticket
ownership, system health) while language models are responsible only for
generating surface prose. Because every significant action emits a structured
\emph{SimEvent} to a persistent log, the resulting corpus and the ground truth
bus are produced by the same run, guaranteeing structural consistency across all
artifact types. This is what distinguishes OrgForge from ``generate fake Slack
messages with an LLM.''

The same architecture serves use cases beyond evaluation: organizational agents
can be run against a live simulation before deployment with verifiable ground
truth to score decisions against; security and compliance tooling requires
cross-system data with known labels that real corpora cannot provide without
legal constraint; and the GraphDynamics subsystem is of independent interest to
organizational behavior researchers studying stress propagation, knowledge
degradation, and communication pattern evolution.
	
	\medskip\noindent
	This paper makes the following contributions:
	\begin{enumerate}[leftmargin=1.5em]
		
		\item \textbf{Formal Simulation Architecture.} A simulation architecture
		formalized as $M = (S, P, V, E)$ that enforces a strict separation between
		fact control and prose generation. This framework prevents LLM
		hallucinations from contaminating synthetic corpora by ensuring that every
		generated document is anchored to deterministic ground truth, a property
		that benefits evaluation, training, and agentic sandbox use cases equally.
		
		\item \textbf{Deterministic Behavioral Logic.} A suite of deterministic
		mechanisms that govern organizational behavior, including graph-dynamic
		processes for stress propagation and automated lifecycle management for
		personas and domain ownership. These mechanisms replace hardcoded heuristics
		with a dynamic, domain-agnostic simulation layer whose behavioral outputs
		are independently useful for organizational research.
		
		\item \textbf{Multi-System Causal Cascades.} A model of integrated
		organizational processes spanning engineering, sales, and support systems.
		The simulation produces cross-system causal cascades in which engineering
		incidents trigger support escalations, CRM state changes, and departmental
		communications, ensuring that generated documents are byproducts of
		realistic, interconnected workflows.
		
		\item \textbf{Open-Source Implementation and Corpus.} A multi-pathway
		knowledge gap detection system paired with an open-source implementation.
		The framework produces fifteen categories of grounded artifacts and employs
		a unified \texttt{SimEvent} bus for verifiable ground truth, enabling
		reproducible organizational corpora for AI development, evaluation, security
		tooling, and organizational simulation research.
		
	\end{enumerate}

\section{Background and Motivation}

\subsection{Corpus Requirements}

Organizational AI systems (whether used for evaluation, training, agentic
deployment, or security tooling) require corpora with at least four properties
that existing resources lack simultaneously:
\begin{enumerate}[leftmargin=1.5em]
	\item \textbf{Traceable ground truth:} each fact must have a canonical
	source that can be used to score retrieval, verify agent decisions, or
	label security events.
	\item \textbf{Temporal structure:} facts must change over time to support
	temporal reasoning, longitudinal training signal, and realistic agent
	environments.
	\item \textbf{Cross-artifact coherence:} the same fact must appear
	consistently across multiple document types.
	\item \textbf{Configurable complexity:} incident severity, organizational
	size, and communication patterns should be tunable.
\end{enumerate}
OrgForge is designed to satisfy all four. Additionally, the framework produces
two properties no existing synthetic dataset provides: \textbf{verified absence}
(silence is ground truth when no simulation state warrants an email) and
\textbf{longitudinal organizational narratives} (the knowledge recovery arc
produces verifiable multi-week stories across all use cases).
	
\subsection{Related Work}

\paragraph{Existing organizational corpora and benchmarks.}
Existing benchmarks, including MultiHop-RAG~\citep{tang2024multihop},
FRAMES~\citep{krishna2024frames}, RAGAS~\citep{es2023ragas}, and
LongBench~\citep{bai2023longbench}, evaluate multi-hop reasoning over static
public corpora but lack cross-system traceability and temporal evolution,
properties required equally for evaluation, training, and agentic deployment.

\paragraph{Multi-hop and cross-document reasoning.}
HotpotQA~\citep{yang2018hotpotqa} and MuSiQue~\citep{trivedi2022musique}
evaluate reasoning over pairs or small sets of Wikipedia passages; QASPER~\citep{dasigi2021qasper}
targets single scientific papers. None contain cross-\emph{system} causal chains where a question about one
artifact (such as why an invoice carries an SLA credit) requires backward
traversal through six distinct subsystems: incident log, Datadog, Zendesk,
Salesforce, email, and JIRA\@. OrgForge corpora are specifically structured
to require this form of causal reasoning, with every link materialized as
a SimEvent.

\paragraph{LLM-based multi-agent simulation.}
Generative Agents~\citep{park2023generativeagents} demonstrated that
LLM-driven agents can produce emergent social behavior in a sandbox town,
and SOTOPIA~\citep{zhou2024sotopia} evaluates social intelligence through
structured agent interactions. These systems treat \emph{behavior} as the
end goal. OrgForge is architecturally distinct: it simulates organizational
\emph{processes} to produce verifiable documentary artifacts, not to evaluate
agent behavior. Critically, in both Generative Agents and SOTOPIA the cognition
layer owns factual state, so cross-document consistency is emergent rather than
guaranteed. OrgForge's validator $V$ makes it a structural guarantee. TheAgentCompany~\citep{xu2025agentcompany} benchmarks LLM agents on
real workplace tasks in a simulated software company, using JIRA, Slack,
and GitHub as the interaction surface. Unlike OrgForge, the environment
is static---agents interact with pre-populated artifacts rather than a
live simulation that evolves causally in response to their decisions.

\paragraph{Synthetic data generation.}
Self-Instruct~\citep{wang2023selfinstruct} and the \textsc{phi-1}
``textbooks are all you need'' methodology~\citep{gunasekar2023phi1}
established that LLM-generated synthetic data can match or exceed the quality
of curated human data for training. Cross-document factual consistency is the
unsolved problem: a hallucinated fact corrupts a training example silently and
an evaluation corpus permanently. Self-consistency filtering partially addresses
this~\citep{es2023ragas} but requires the generating model to arbitrate its own
errors. OrgForge externalizes truth entirely: the LLM renders facts into prose
but never controls them.

\paragraph{Factual consistency and hallucination.}
Hallucination in abstractive generation has been characterized along
faithfulness and factuality dimensions~\citep{maynez2020faithfulness}, and
benchmark-level factual consistency has been quantified via learned
metrics~\citep{kryscinski2020factcc}. FEVER~\citep{thorne2018fever} frames
fact verification as a retrieval-and-classification task over Wikipedia claims.
These approaches treat hallucination as a property to detect; OrgForge prevents contamination architecturally.

\paragraph{Corporate and enterprise corpora.}
The Enron corpus~\citep{klimt2004enron} remains the canonical corporate email
dataset but is now two decades old and represents a singular, pathological
organizational context. The Avocado Research Email Collection~\citep{oard2015avocado}
provides a more recent corporate email archive but, like Enron, consists of
email alone with no associated ticketing, incident, CRM, or financial records.

\paragraph{Agent-based organizational simulation.}
Agent-based modeling has been applied to organizational behavior since
Carley's work~\citep{carley2002simulating}. Frameworks such as
Mesa~\citep{masad2015mesa} provide general-purpose simulation infrastructure.
Social network simulations study information
diffusion~\citep{watts1998smallworld}. None connect organizational simulation
to document corpus generation. OrgForge bridges this gap.
	
	\section{System Architecture}
	\label{sec:architecture}
	
	OrgForge runs a discrete-time simulation over $N$ days. Each day proceeds through a planning phase, an execution phase, and an end-of-day summarization phase. The core invariant is that the SimEvent log is the sole authoritative record of facts; all generated text is prose grounded in that record.
	
	\begin{figure}[H]
		\centering
		\begin{tikzpicture}[
			node distance=0.38cm and 2.4cm,
			phase/.style={
				rectangle, rounded corners=5pt, draw, thick,
				minimum width=3.6cm, minimum height=0.72cm,
				font=\small\bfseries\sffamily, align=center,
				inner xsep=8pt, inner ysep=5pt
			},
			pstep/.style={
				rectangle, rounded corners=3pt, draw=gray!45,
				fill=white,
				minimum width=3.2cm, minimum height=0.56cm,
				font=\scriptsize\sffamily, align=center,
				inner xsep=6pt
			},
			emit/.style={
				rectangle, rounded corners=2pt,
				fill=gray!10, draw=gray!35,
				font=\tiny\ttfamily, align=center,
				inner xsep=4pt, inner ysep=2pt
			},
			arrow/.style={-{Stealth[length=4pt]}, thick, gray!55},
			earrow/.style={-{Stealth[length=3pt]}, thin, gray!35},
			looplabel/.style={font=\scriptsize\itshape, text=gray!55},
			]
			
			
			\node[phase, fill=teal!8, draw=teal!50, text=teal!80!black]
			(plan) at (0, 0) {Planning Phase};
			
			\node[phase, fill=violet!7, draw=violet!45, text=violet!75!black,
			right=2.6cm of plan]
			(exec) {Execution Phase};
			
			\node[phase, fill=orange!8, draw=orange!45, text=orange!75!black,
			right=2.6cm of exec]
			(eod) {End-of-Day Phase};
			
			
			\node[pstep, fill=teal!5, draw=teal!35, below=0.45cm of plan]
			(p1) {Capacity computation ($C_i$)};
			\node[pstep, fill=teal!5, draw=teal!35, below=0.32cm of p1]
			(p2) {Dept planners observe $S$};
			\node[pstep, fill=teal!5, draw=teal!35, below=0.32cm of p2]
			(p3) {Validator $V$ admits/rejects};
			\node[pstep, fill=teal!5, draw=teal!35, below=0.32cm of p3]
			(p4) {Hungarian ticket assignment};
			
			\node[emit, right=0.22cm of p3.east, anchor=west]
			(ve) {\texttt{plan\_validated}};
			
			
			\node[pstep, fill=violet!5, draw=violet!35,
			below=0.45cm of exec]
			(e1) {Agenda dispatch per actor};
			\node[pstep, fill=violet!5, draw=violet!35, below=0.32cm of e1]
			(e2) {Artifact generators run};
			\node[pstep, fill=violet!5, draw=violet!35, below=0.32cm of e2]
			(e3) {SimClock \texttt{advance\_actor}};
			\node[pstep, fill=violet!5, draw=violet!35, below=0.32cm of e3]
			(e4) {Signal-driven emails fire};
			
			\node[emit, right=0.22cm of e2.east, anchor=west]
			(ae) {\texttt{artifact\_created}};
			\node[emit, right=0.22cm of e4.east, anchor=west]
			(ee) {\texttt{inbound\_external\_email}};
			
			
			\node[pstep, fill=orange!5, draw=orange!35,
			below=0.45cm of eod]
			(d1) {GraphDynamics tick};
			\node[pstep, fill=orange!5, draw=orange!35, below=0.32cm of d1]
			(d2) {Morale / stress update};
			\node[pstep, fill=orange!5, draw=orange!35, below=0.32cm of d2]
			(d3) {CRM edge-weight sync};
			\node[pstep, fill=orange!5, draw=orange!35, below=0.32cm of d3]
			(d4) {\texttt{day\_summary} emitted};
			
			\node[emit, right=0.22cm of d4.east, anchor=west]
			(de) {\texttt{day\_summary}};
			
			
			\foreach \a/\b in {p1/p2, p2/p3, p3/p4}
			\draw[arrow] (\a.south) -- (\b.north);
			\foreach \a/\b in {e1/e2, e2/e3, e3/e4}
			\draw[arrow] (\a.south) -- (\b.north);
			\foreach \a/\b in {d1/d2, d2/d3, d3/d4}
			\draw[arrow] (\a.south) -- (\b.north);
			
			
			\draw[earrow] (p3.east) -- (ve.west);
			\draw[earrow] (e2.east) -- (ae.west);
			\draw[earrow] (e4.east) -- (ee.west);
			\draw[earrow] (d4.east) -- (de.west);
			
			
			\draw[arrow, teal!60]   (plan.east)  -- (exec.west)
			node[midway, above, font=\scriptsize\itshape, text=gray!55] {proposals};
			\draw[arrow, violet!55] (exec.east)  -- (eod.west)
			node[midway, above, font=\scriptsize\itshape, text=gray!55] {events};
			
			
			\coordinate (loopstart) at ([yshift=-0.45cm]d4.south);
			\coordinate (loopend)   at ([yshift=-0.45cm]p1.south);
			
			\begin{pgfonlayer}{background}
				\draw[arrow, gray!40, rounded corners=6pt]
				([yshift=-0.3cm]d4.south)
				-- ++(0,-0.35cm)
				-| ([xshift=-0.6cm]loopend)
				-- ([xshift=-0.6cm]p1.south |- loopend)
				-- ([xshift=-0.6cm]p1.south)
				-- (p1.south);
			\end{pgfonlayer}
			
			\node[looplabel] at
			($(loopend)!0.5!(loopstart) + (0,-0.22cm)$)
			{next day $t+1$};
			
		\end{tikzpicture}
		\caption{Day lifecycle in OrgForge. Each simulated day executes three phases.
			The Planning Phase computes actor capacities, collects department proposals,
			validates them via $V$, and locks ticket assignments. The Execution Phase
			dispatches artifact generators and advances per-actor simulation clocks.
			The End-of-Day Phase ticks GraphDynamics, updates morale and stress, and
			synchronizes CRM edge weights before emitting a \texttt{day\_summary} SimEvent.
			Pill-shaped labels denote SimEvents emitted to the ground truth bus at each step.}
		\label{fig:day-lifecycle}
	\end{figure}
	
	\subsection{Formal System Definition}
	
	We define OrgForge as the tuple $M = (S, P, V, E)$:
	
	\begin{itemize}[leftmargin=1.5em]
		\item $\mathbf{S}$ \textbf{(State):} All mutable simulation variables: system health $H \in [0, 100]$, team morale $\mu \in [0, 1]$, active incidents, tickets, Confluence registry, per-engineer stress, CRM state, and the DomainRegistry.
		
		\item $\mathbf{P}$ \textbf{(Planners):} LLM-based department agents that observe $S$ and the SimEvent history, then generate structured JSON proposals. Planners influence narrative direction but cannot mutate $S$ or write to $E$ directly.
		
		\item $\mathbf{V}$ \textbf{(Validator):} A deterministic function $V: \text{ProposedEvent} \times S \times E \to \{0, 1\}$ that admits or rejects each proposal before execution.
		
		\item $\mathbf{E}$ \textbf{(Events):} The SimEvent log: a persistent, append-only record of every significant action. This is the ground truth bus.
	\end{itemize}
	
	The boundary $V$ separates the ``physics'' layer ($S$, $E$, graph dynamics, CRM, DomainRegistry) from the ``cognition'' layer ($P$). This prevents hallucinations from contaminating the corpus.
	
	\begin{figure}[H]
		\centering
		\begin{tikzpicture}[
			node distance=0.55cm and 1.8cm,
			inner box/.style={rectangle, rounded corners=3pt, draw=gray!50,
				fill=white, minimum width=3.4cm, minimum height=0.55cm,
				font=\footnotesize, align=center},
			container/.style={rectangle, rounded corners=6pt, draw, thick,
				inner sep=0.3cm},
			arrow/.style={-{Stealth[length=4pt]}, thick},
			]
			
			\node[inner box, fill=teal!8,  draw=teal!40]  (state1) {System health / team morale};
			\node[inner box, fill=teal!8,  draw=teal!40, below=0.22cm of state1] (state2) {Per-actor stress + CRM state};
			\node[inner box, fill=teal!8,  draw=teal!40, below=0.22cm of state2] (state3) {SimEvent ground truth bus};
			\node[inner box, fill=teal!8,  draw=teal!40, below=0.22cm of state3] (state4) {Social graph + DomainRegistry};
			\node[inner box, fill=teal!8,  draw=teal!40, below=0.22cm of state4] (state5) {Ticket assignment (Hungarian)};
			
			\begin{scope}[on background layer]
				\node[container, fill=teal!5, draw=teal!60,
				fit=(state1)(state2)(state3)(state4)(state5), inner sep=0.35cm] (leftbox) {};
			\end{scope}
			\node[font=\small\bfseries, text=teal!80!black, anchor=south] at (leftbox.north) {\strut Deterministic engine (Python)};
			\node[font=\footnotesize\itshape, text=teal!60, anchor=north] at (leftbox.south) {\strut controls all factual state};
			
			\node[inner box, fill=violet!7, draw=violet!35, right=6cm of state1] (llm1) {Slack messages};
			\node[inner box, fill=violet!7, draw=violet!35] at (state2 -| llm1) (llm2) {PR descriptions / emails};
			\node[inner box, fill=violet!7, draw=violet!35] at (state3 -| llm1) (llm3) {Confluence pages};
			\node[inner box, fill=violet!7, draw=violet!35] at (state4 -| llm1) (llm4) {Zoom transcripts};
			\node[inner box, fill=violet!7, draw=violet!35] at (state5 -| llm1) (llm5) {Customer / vendor prose};
			
			\begin{scope}[on background layer]
				\node[container, fill=violet!4, draw=violet!50,
				fit=(llm1)(llm2)(llm3)(llm4)(llm5), inner sep=0.35cm] (rightbox) {};
			\end{scope}
			\node[font=\small\bfseries, text=violet!70!black, anchor=south] at (rightbox.north) {\strut LLM generation layer};
			\node[font=\footnotesize\itshape, text=violet!55, anchor=north] at (rightbox.south) {\strut renders surface prose only};
			
			\draw[arrow, gray!60]
			($(leftbox.north east)!0.25!(leftbox.south east)$) --
			node[above, font=\scriptsize\itshape, text=gray!55] {validated proposals}
			($(rightbox.north west)!0.25!(rightbox.south west)$);
			\draw[arrow, gray!60]
			($(leftbox.north east)!0.50!(leftbox.south east)$) --
			node[above, font=\scriptsize\itshape, text=gray!55] {injection points}
			($(rightbox.north west)!0.50!(rightbox.south west)$);
			\draw[arrow, gray!60]
			($(leftbox.north east)!0.75!(leftbox.south east)$) --
			node[above, font=\scriptsize\itshape, text=gray!55] {constrained prompts}
			($(rightbox.north west)!0.75!(rightbox.south west)$);
			
			\draw[dashed, gray!40, thick]
			($(leftbox.north east)!0.5!(rightbox.north west)$) --
			($(leftbox.south east)!0.5!(rightbox.south west)$);
			\node[font=\footnotesize\itshape, text=gray!55, fill=white, inner sep=2pt]
			at ($(leftbox.north east)!0.5!(rightbox.north west) + (0, 0.4cm)$)
			{physics-cognition boundary};
		\end{tikzpicture}
		\caption{Architecture of the physics-cognition boundary. The deterministic engine owns all factual state including CRM records, the DomainRegistry, and ticket assignments; LLMs generate surface prose from validated proposals.}
		\label{fig:architecture}
	\end{figure}
	
	\subsection{Prompt-Level Fact Grounding}
	\label{sec:fact-grounding}
	
	The physics-cognition boundary described above is enforced at three
	layers, each of which independently prevents LLM-generated prose from
	diverging from the SimEvent record.
	
	\paragraph{Layer 1: Fact injection into every prompt.}
	Every artifact generator receives the specific SimEvent-level facts as
	prompt context before the LLM runs. Ticket-progress prompts contain the
	exact ticket ID, title, status, and recent comments. PR-review prompts
	contain the PR title, author, linked ticket, recurrence history, and
	reviewer expertise vector. Incident-summary prompts contain the root
	cause string, escalation narrative, and affected tech-stack components.
	External-contact prompts contain the incident ID, root cause, and the
	contact's derived tone. Standup prompts contain the explicit owned-ticket
	list with the instruction to reference only those tickets. The LLM does
	not choose \emph{what} to write about; the engine locks the topic before
	generation begins.
	
	\paragraph{Layer 2: Structured JSON output for state transitions.}
	All state-affecting decisions are parsed from structured JSON fields, not
	from prose. The ticket-progress handler parses
	\texttt{is\_code\_complete} (boolean) to decide whether to spawn a PR\@.
	The PR-review handler parses \texttt{verdict}
	$\in \{\texttt{approved},\, \texttt{changes\_requested}\}$ to decide
	merge versus revision. The async-thread classifier parses
	\texttt{outcome} to classify knowledge gaps. If the LLM writes a
	positive review in prose but sets \texttt{verdict:
		"changes\_requested"}, the engine requests changes. Prose and state
	transitions are decoupled by construction.
	
	\paragraph{Layer 3: SimEvents are emitted by the engine, not derived
		from prose.}
	Each SimEvent is constructed by the Python handler using engine-owned
	facts; the generated prose is stored as a separate artifact. The
	SimEvent payload contains the canonical fact record (incident ID, root
	cause, ticket status, CRM stage); the prose artifact is the retrieval
	surface. No downstream state transition reads from prose, and no
	SimEvent field is populated by LLM output.
	
	\noindent Together, these three layers mean that the physics-cognition
	boundary is not a single architectural wall but a defense-in-depth
	stack: topics are locked before generation, state transitions are parsed
	from structured fields, and the ground truth bus is written exclusively
	by the deterministic engine.
	The 0.9962 Prose-SimEvent fidelity score
	(Section~\ref{sec:consistency-eval}) empirically validates that LLMs
	surface the injected facts at high rates; the physics-cognition boundary
	ensures correctness is independent of that rate.
	
	\subsection{The SimEvent Ground Truth Bus}
	
	Every significant action emits a \texttt{SimEvent}: a structured record persisted to MongoDB with a timestamp, event type, actor set, and payload. SimEvents are the canonical source of truth, capturing: \texttt{event\_type}, \texttt{actors}, \texttt{artifact\_ids} (JIRA tickets, PR numbers, Confluence pages, ZD tickets, SF opportunities), and \texttt{facts} (key-value payload at the moment of emission). Each day also emits a \texttt{day\_summary} SimEvent providing a queryable temporal index.
	
	\subsection{The Social Graph and Initial Conditions}
	
	The simulation maintains a weighted undirected graph $G = (V_G, E_G)$ where nodes are employees and external contacts, and edge weights $w_{uv}$ represent relationship strength:
	
	\begin{equation}
		w_{uv}^{(0)} =
		\begin{cases}
			10.5 & \text{if } u, v \text{ are in the same department} \\
			5.5  & \text{if } u, v \text{ are both department leads} \\
			0.5  & \text{otherwise (cross-department or external)}
		\end{cases}
		\label{eq:initial-weights}
	\end{equation}
	
	\subsection{GraphDynamics: Formal Specification}
	\label{sec:graphdynamics}
	
	\subsubsection{A. Stress Propagation via Betweenness Centrality}
	
	Let $S_{i,t}$ denote the stress of agent $i$ at the end of day $t$, and $g(i)$ the betweenness centrality of node $i$. Key players are defined as:
	
	\begin{equation}
		\KP_t = \bigl\{\, i \in V_G \;\big|\; g(i) \;\geq\; \tilde{g}_t \cdot \lambda \,\bigr\}
		\label{eq:key-players}
	\end{equation}
	
	where $\tilde{g}_t$ is the median betweenness centrality and $\lambda = 2.0$. The stress update is:
	
	\begin{equation}
		S_{i,\,t+1} = \max\!\left(0,\; S_{i,t} - \delta\right)
		+ \sum_{j \in \KP_t} \indicator{S_{j,t} > B}
		\cdot (S_{j,t} - B) \cdot \alpha \cdot
		\frac{w_{ji}}{\displaystyle\sum_{k \in \Neigh(j)} w_{jk}}
		\label{eq:stress-propagation}
	\end{equation}
	
	where $\delta = 3$ (daily recovery), $B = 65$ (burnout threshold), $\alpha = 0.25$ (bleed rate).
	
	\subsubsection{B. Temporal Edge-Weight Decay}
	
	\begin{equation}
		w_{uv,\,t+1} = \max\!\left(\omega_{\min},\;
		w_{uv,\,t} \cdot \gamma + \beta\!\left(I_{uv,t}\right)
		\right)
		\label{eq:edge-decay}
	\end{equation}
	
where $\gamma = 0.97$, $\omega_{\min} = 0.5$,
$I_{uv,t} \in \{\textsc{inc},\, \textsc{pr},\, \textsc{1on1},\,
\textsc{slack},\, \varnothing\}$ is the highest-priority interaction
type between $u$ and $v$ on day $t$, and $\beta$ is the corresponding
interaction boost: $\beta_{\textsc{inc}} = 4.0$ (joint incident),
$\beta_{\textsc{pr}} = 3.0$ (PR co-review),
$\beta_{\textsc{1on1}} = 2.0$, $\beta_{\textsc{slack}} = 1.5$,
$\beta_{\varnothing} = 0$.
	
	\subsubsection{C. Shortest-Path Escalation Routing}
	
	Escalation is modeled as shortest-path on an inverse-weight graph with cost $w'_{uv} = 1/\max(w_{uv}, 0.01)$:
	
	\begin{equation}
		L(u, v^*) = \argmin_{\text{path } P_{uv^*}}
		\sum_{(e_1,\, e_2) \in P_{uv^*}} w'_{e_1 e_2}
		\label{eq:escalation-path}
	\end{equation}
	
	computed via Dijkstra's algorithm. The escalation chain is emitted as a SimEvent and fed to the LLM as prompt context.
	
	\subsubsection{D. CRM-Driven Edge-Weight Synchronization and Stress Feedback}
	\label{sec:crm-graph-feedback}
	
	A fourth mechanism closes the loop between the CRM state machine
	(Section~\ref{sec:crm}) and GraphDynamics, creating a bidirectional coupling
	that the one-directional description in Section~\ref{sec:crm} does not capture.
	At end-of-day, \texttt{sync\_crm\_edge\_weights()} adjusts the weight of every
	edge connecting an external contact node to its internal liaison:
	
	\begin{equation}
		w_{uv,\,t+1}^{\scriptscriptstyle\text{CRM}} =
		\max\!\left(\omega_{\min},\;
		w_{uv,\,t}
		+ \Delta_{\textsc{zd}}(u)
		+ \Delta_{\textsc{opp}}(u)
		- \rho_{\textsc{risk}}(u)
		- \rho_{\textsc{silence}}(u)
		\right)
		\label{eq:crm-edge-sync}
	\end{equation}
	
	where the boosts and penalties are:
	
	\begin{align*}
		\Delta_{\textsc{zd}}(u)        &= 2.0 \cdot \indicator{\text{open ZD}} + 1.0 \cdot \indicator{\text{Urgent ZD}} \\
		\Delta_{\textsc{opp}}(u)       &= 3.0 \cdot \indicator{\text{open opp}} + 2.0 \cdot \indicator{\text{Negotiation/Review}} \\
		\rho_{\textsc{risk}}(u)        &= 2.5 \cdot \indicator{\text{risk\_notes}} + 1.0 \cdot \indicator{\text{risk\_flag}} \\
		\rho_{\textsc{silence}}(u)     &= 0.5 \cdot \indicator{\text{no opp} \wedge \text{no ticket}}
	\end{align*}
	
	This ensures that Dijkstra escalation routing (Equation~\ref{eq:escalation-path})
	naturally favours the account liaison for an at-risk customer, and that
	liaison edges with dormant accounts decay toward estrangement over time.
		
	\subsection{The Proposal-Validation Loop}
	\label{sec:validator}
	
	The validator implements $V$ via five ordered checks: (1)~actor integrity:
	every actor must exist in \texttt{org\_chart} or \texttt{external\_contacts};
	(2)~novel event triage: unknown event types approved only with known
	\texttt{artifact\_hint}; (3)~state plausibility: celebrations blocked when
	$H < 40$; (4)~cooldown windows: minimum inter-event gaps; (5)~morale gating:
	\texttt{morale\_intervention} only when $\mu \leq 0.6$.
	
	\subsection{Multi-Department Planning}
	
	Each day begins with a \texttt{DayPlannerOrchestrator}. Engineering plans
	first as the primary driver. Other departments (Sales, HR, Product, Design,
	QA) plan reactively, receiving Engineering's plan as input. An
	\texttt{OrgCoordinator} identifies cross-department collision events.
	
	CRM state is injected into every department planner via \texttt{crm.planner\_context()}, which returns a compact summary of open support tickets and at-risk deals. The coordinator uses this signal to seed realistic Sales/Support$\,\leftrightarrow\,$Engineering collisions.
	
	Engineer capacity is computed dynamically:
	
	\begin{equation}
		C_i = \max\!\left(1.5,\; 6.0
		- 1.5 \cdot \indicator{i \in \textsc{on-call}}
		- 2.0 \cdot \indicator{S_{i,t} > 80}
		- 1.0 \cdot \indicator{60 < S_{i,t} \leq 80}
		\right)
		\label{eq:capacity}
	\end{equation}
	
	\paragraph{Temporal grounding.}
	Every department planning prompt contains the explicit instruction that the
	simulation start day is the first day the corpus \emph{observes} the
	organization, not its founding day, and that years of existing code, legacy
	systems, and established processes already exist. This prevents the LLM from
	generating greenfield artifacts on Day~1 and produces output consistent with
	a mature organization.
	
	\subsection{Embedding-Based Ticket Assignment}
	\label{sec:ticket-assignment}
	
	The \texttt{TicketAssigner} replaces hardcoded skill-keyword matching with cosine similarity between engineer expertise embeddings and ticket title embeddings. This is the mechanism that makes the simulation domain-agnostic at the assignment layer---the same code works for any industry defined in \texttt{config.yaml}.
	
	For each (engineer, ticket) pair, the composite score is:
	
	\begin{equation}
		c_{ij} = \text{sim}(\mathbf{e}_i, \mathbf{t}_j) \cdot \left(1 - \frac{S_i}{100}\right) \cdot \left(1 - 0.3 \cdot g(i)\right) \cdot r_{ij}
		\label{eq:assignment-score}
	\end{equation}
	
	where $\text{sim}(\mathbf{e}_i, \mathbf{t}_j)$ is cosine similarity between the engineer's expertise vector and the ticket title vector, rescaled to $[0.5, 1.5]$; $S_i/100$ is inverse stress; $g(i)$ is betweenness centrality; and $r_{ij} = 1.2$ if the engineer has prior history with this ticket, $1.0$ otherwise.
	
	The globally optimal assignment is obtained via the Hungarian algorithm~\citep{kuhn1955hungarian}:
	
	\begin{equation}
		\sigma^* = \argmin_{\sigma \in \Pi} \sum_{i} \left(-c_{i,\sigma(i)}\right)
		\label{eq:hungarian}
	\end{equation}
	
	where $\Pi$ is the set of feasible one-to-one matchings respecting capacity constraints. All embeddings are produced by Qwen3-Embedding-4B~\citep{zhang2025qwen3embedding}
	($d = 2{,}560$).\footnote{\url{https://huggingface.co/Qwen/Qwen3-Embedding-4B}}
	Engineer expertise vectors are computed once at genesis and stored in MongoDB;
	ticket title vectors are computed on creation and cached at the same time.
	Ownership is locked before any LLM planning runs, so ownership conflicts are
	structurally impossible.

	\subsection{Non-Engineering Department Simulation}
	\label{sec:non-eng}
	
	The simulation produces full work cycles for all departments. The \texttt{dept\_type} and \texttt{completion\_artifact} fields stamped at sprint planning time control routing:
	
	\begin{itemize}[leftmargin=1.5em]
		\item \textbf{Sales} completes tickets via customer-facing outbound emails tied to Salesforce opportunities.
		\item \textbf{HR} sends offer letters and onboarding prep emails to incoming hires.
		\item \textbf{Product} creates JIRA tickets from customer complaints via a Sales$\to$Product triage pipeline.
		\item \textbf{Design} and \textbf{QA} complete tickets via Confluence pages or Slack threads.
	\end{itemize}
	
	Non-engineering tickets \emph{never} produce PRs. Each department's completion artifacts have full causal chain parity with engineering artifacts, so cross-department evaluation queries work uniformly.
	
	\subsection{Morale Dynamics and Sentiment Feedback}
	
	Team morale $\mu_t$ evolves via multiplicative decay with conditional recovery:
	\begin{equation}
		\mu_{t+1} =
		\begin{cases}
			\min\!\left(1.0,\;\mu_t \cdot \gamma_\mu + \rho\right) & \text{if no active incidents at EOD} \\
			\mu_t \cdot \gamma_\mu & \text{otherwise}
		\end{cases}
	\end{equation}
	
	where $\gamma_\mu = 0.98$ (daily decay) and $\rho = 0.05$ (recovery increment).
	
	VADER sentiment scoring~\citep{hutto2014vader} on Slack artifacts provides
	bounded feedback: negative sentiment increments stress by up to $+5$;
	positive provides up to $-5$ recovery.
	
	\subsection{Artifact Generation}
	\label{sec:artifacts}
	
	Once the day plan is validated, \texttt{NormalDayHandler} dispatches each engineer's agenda items to typed artifact generators. Table~\ref{tab:artifacts} summarizes the complete output inventory.
	
\begin{longtable}{p{2.8cm}p{5.2cm}p{4.8cm}}
	\toprule
	\textbf{Category} & \textbf{Artifacts} & \textbf{SimEvent types} \\
	\midrule
	\endfirsthead
	
	\toprule
	\textbf{Category} & \textbf{Artifacts} & \textbf{SimEvent types} \\
	\midrule
	\endhead
	
	\midrule
	\multicolumn{3}{r}{\small\itshape continued on next page} \\
	\endfoot
	
	\bottomrule
	\addlinespace[0.5em]
	\caption{Complete artifact output inventory. Runtime artifacts are produced
		during simulation; post-simulation artifacts are derived from the SimEvent
		log without re-running the simulation.}
	\label{tab:artifacts}
	\endlastfoot
	
	\multicolumn{3}{l}{\textit{Internal coordination}} \\
	\quad Slack & Dept channels, DMs, digital-hq, random &
	\texttt{async\_question}, \texttt{1on1}, \texttt{org\_collision},
	\texttt{watercooler\_chat} \\
	\quad JIRA & Tickets + per-comment files &
	\texttt{ticket\_progress}, \texttt{jira\_ticket\_created} \\
	\quad GitHub & PRs with review comments & \texttt{pr\_review} \\
	\quad Confluence & Genesis, postmortems, design docs, ad-hoc &
	\texttt{confluence\_created} \\
	\quad Zoom & Timestamped meeting transcripts (.md) &
	\texttt{design\_discussion} \\
	\midrule
	
	\multicolumn{3}{l}{\textit{Email}} \\
	\quad Inbound & Vendor alerts, customer complaints/questions/feature requests &
	\texttt{inbound\_external\_email} \\
	\quad Outbound & Sales outreach, customer replies, vendor acks, HR correspondence &
	\texttt{sales\_outbound\_email}, \texttt{customer\_reply\_sent},
	\texttt{hr\_outbound\_email} \\
	\midrule
	
	\multicolumn{3}{l}{\textit{CRM / customer-facing}} \\
	\quad Zendesk & Tickets + per-comment files &
	\texttt{zd\_ticket\_opened}, \texttt{zd\_tickets\_escalated} \\
	\quad Salesforce & Accounts + opportunities &
	\texttt{crm\_touchpoint}, \texttt{sf\_deals\_risk\_flagged} \\
	\midrule
	
	\multicolumn{3}{l}{\textit{Post-simulation derived}} \\
	\quad NPS & Survey responses + summary & Derived from SimEvent log \\
	\quad Invoices & SLA-adjusted per-customer & Derived from incident duration \\
	\quad Datadog & Metrics (.jsonl, 15-min) + alerts & Derived from health timeseries \\
	\midrule
	
	\multicolumn{3}{l}{\textit{Ground truth / debugging}} \\
	\quad SimEvent log & MongoDB + exportable & All types \\
	\quad DomainRegistry & Live mutable state & \texttt{domain\_ownership\_claimed} \\
	\quad Assignment scores & Per-sprint scoring matrix & (debugging collection) \\
	\midrule
	
	\multicolumn{3}{l}{\textit{Security telemetry~\citep{flynt2026orgforgeit}}} \\
	\quad DLP \& IDP logs & Access logs (JSONL / CEF / ECS / LEEF) + ground truth &
	\texttt{dlp\_alert}, \texttt{secret\_detected} \\
	
\end{longtable}
	
	\paragraph{Meeting medium routing and retrieval difficulty gradient.}
	Design discussions route to Zoom when the participant count exceeds one,
	the topic is architectural or cross-cutting, and team morale is not
	critically low. Zoom transcripts are saved as timestamped Markdown,
	representing decisions made verbally that never surface in Confluence or
	JIRA unless someone writes them up. The probability-gated Confluence
	spawn (Equation~\ref{eq:async-doc-prob} below, $P = 0.30$ for escalated
	threads) means approximately 70\% of Zoom-originated decisions exist
	\emph{only} in the transcript. This produces a natural retrieval
	difficulty gradient: easy questions pull from Confluence pages indexed by
	title and system tag; hard questions require retrieval from Zoom
	transcripts where the same fact is embedded in conversational turns
	without structured metadata. The gradient is a deliberate corpus design
	property, not an artifact of incomplete generation.
	
	\paragraph{Expertise-matched participant selection.}
	Async questions and design discussions select participants via
	\texttt{\_expertise\_matched\_participants()}, which computes cosine
	similarity between the topic embedding and each candidate's expertise
	vector, weighted by social-graph edge weight to the initiator. This
	ensures that technical threads attract domain-relevant participants
	while remaining socially plausible---a guard against the common
	multi-agent simulation failure mode of off-domain actors joining
	every conversation.
	
	\paragraph{Watercooler chat as structured noise.}
	Each engineer has a configurable per-day probability of initiating a
	non-work chat. Topics are derived from shared participant interests,
	current stress levels, and time of day. These threads create realistic
	off-topic noise in the corpus: the kind of Slack messages that
	production RAG systems must learn to filter or deprioritize. Because
	watercooler threads emit \texttt{watercooler\_chat} SimEvents, their
	presence is labeled ground truth, enabling evaluation of retrieval
	precision under realistic noise conditions.

	\subsection{Dynamic Persona Injection and Voice Gating}
	
	A context-aware persona injection mechanism prevents linguistic homogenization. The voice-card function $V(i, \text{ctx}) = \mathcal{F}(S_{i,t}, \mathcal{P}_i, \text{ctx})$ maps stress, persona, and context to natural-language prompting constraints. Key properties:
	
	\begin{itemize}[leftmargin=1.5em]
		\item \textbf{State-to-mood mapping:} stress $S_{i,t} > 80$ renders ``visibly stressed and terse''; $S_{i,t} < 60$ renders ``relaxed and present.''
		\item \textbf{Contextual field gating:} interests injected only during \texttt{watercooler}; anti-patterns only in high-friction contexts.
		\item \textbf{CRM pressure injection:} at-risk deals and urgent tickets are injected into voice cards via \texttt{crm\_pressure\_hint()}.
		\item \textbf{External contact voice cards:} vendors and customers get persona generation from \texttt{inbound\_email\_sources} with sentiment-derived mood.
	\end{itemize}
	
	\subsection{Causal Chain Tracking and Recurrence Detection}
	\label{sec:causalchain}
	
	Once an incident opens, a \texttt{CausalChainHandler} accumulates an ordered list of artifact IDs. Snapshots are written into each SimEvent's \texttt{facts} payload. Recurrence detection fuses vector and text search via Reciprocal Rank Fusion:
	
	\begin{equation}
		s_{\textsc{rrf}}(d) =
		0.65 \cdot \frac{1}{60 + r_{\textsc{vec}}(d)}
		+ 0.35 \cdot \frac{1}{60 + r_{\textsc{txt}}(d)}
	\end{equation}
	
	\subsection{Signal-Driven Customer Email Generation}
	\label{sec:email}
	
	Customer emails are entirely signal-driven. The function \texttt{\_derive\_customer\_email\_signals()} inspects live simulation state and fires emails only when warranted. Let $\mathcal{C}$ be the set of customer sources. For each $c \in \mathcal{C}$, the engine evaluates a priority-ordered signal cascade:
	
	\begin{equation}
		\text{signal}(c, t) =
		\begin{cases}
			\texttt{complaint}       & \exists\, \text{incident affecting } c.\text{depends\_on} \\
			\texttt{question}        & \exists\, \text{opp at Negotiation stale} > 3\text{d} \\
			\texttt{question}        & c.\text{renewal\_date} - t \leq 60\text{d} \\
			\texttt{complaint/question} & \exists\, \text{risky opp} \wedge c.\text{sentiment} < 0.6 \\
			\texttt{feature\_request} & c.\text{expansion} \geq 8 \wedge H \geq 80 \quad (p = 0.25) \\
			\texttt{complaint}       & c.\text{sentiment} < 0.45 \quad (p = 0.15) \\
			\varnothing              & \text{otherwise (silence is correct)}
		\end{cases}
		\label{eq:email-signals}
	\end{equation}
	
\begin{figure}[!t]
	\centering
	\small
	\begin{tikzpicture}[row/.style={font=\scriptsize\sffamily}]
		
		\def\cw{6.2cm}
		\def\sw{2.8cm}
		\def\tw{3.2cm}
		\def\rh{0.62cm}
		
		\node[rectangle, fill=gray!12, draw=gray!40,
		minimum width=\cw, minimum height=\rh,
		font=\scriptsize\bfseries\sffamily, align=center, anchor=north west]
		(h1) at (0,0) {Condition evaluated};
		\node[rectangle, fill=gray!12, draw=gray!40,
		minimum width=\sw, minimum height=\rh,
		font=\scriptsize\bfseries\sffamily, align=center, anchor=north west]
		at (\cw,0) {Email fired};
		\node[rectangle, fill=gray!12, draw=gray!40,
		minimum width=\tw, minimum height=\rh,
		font=\scriptsize\bfseries\sffamily, align=center, anchor=north west]
		at (\cw+\sw,0) {Type};
		
		\node[rectangle, fill=red!6, draw=gray!30, minimum width=\cw, minimum height=\rh,
		font=\scriptsize\sffamily, align=left, inner xsep=6pt, anchor=north west]
		at (0,-\rh) {Active incident affects \texttt{depends\_on\_components}};
		\node[rectangle, fill=red!6, draw=gray!30, minimum width=\sw, minimum height=\rh,
		font=\scriptsize\sffamily, align=center, anchor=north west]
		at (\cw,-\rh) {yes};
		\node[rectangle, fill=red!8, draw=gray!30, minimum width=\tw, minimum height=\rh,
		font=\scriptsize\ttfamily, align=center, anchor=north west]
		at (\cw+\sw,-\rh) {complaint};
		
		\node[rectangle, fill=blue!4, draw=gray!30, minimum width=\cw, minimum height=\rh,
		font=\scriptsize\sffamily, align=left, inner xsep=6pt, anchor=north west]
		at (0,-2*\rh) {SF opp at Negotiation stage, stale $>$ 3 days};
		\node[rectangle, fill=blue!4, draw=gray!30, minimum width=\sw, minimum height=\rh,
		font=\scriptsize\sffamily, align=center, anchor=north west]
		at (\cw,-2*\rh) {yes};
		\node[rectangle, fill=blue!6, draw=gray!30, minimum width=\tw, minimum height=\rh,
		font=\scriptsize\ttfamily, align=center, anchor=north west]
		at (\cw+\sw,-2*\rh) {question};
		
		\node[rectangle, fill=blue!4, draw=gray!30, minimum width=\cw, minimum height=\rh,
		font=\scriptsize\sffamily, align=left, inner xsep=6pt, anchor=north west]
		at (0,-3*\rh) {Renewal date within 60 days};
		\node[rectangle, fill=blue!4, draw=gray!30, minimum width=\sw, minimum height=\rh,
		font=\scriptsize\sffamily, align=center, anchor=north west]
		at (\cw,-3*\rh) {yes};
		\node[rectangle, fill=blue!6, draw=gray!30, minimum width=\tw, minimum height=\rh,
		font=\scriptsize\ttfamily, align=center, anchor=north west]
		at (\cw+\sw,-3*\rh) {question};
		
		\node[rectangle, fill=orange!5, draw=gray!30, minimum width=\cw, minimum height=\rh,
		font=\scriptsize\sffamily, align=left, inner xsep=6pt, anchor=north west]
		at (0,-4*\rh) {Risky opp $\wedge$ customer sentiment $< 0.6$};
		\node[rectangle, fill=orange!5, draw=gray!30, minimum width=\sw, minimum height=\rh,
		font=\scriptsize\sffamily, align=center, anchor=north west]
		at (\cw,-4*\rh) {yes};
		\node[rectangle, fill=orange!8, draw=gray!30, minimum width=\tw, minimum height=\rh,
		font=\scriptsize\ttfamily, align=center, anchor=north west]
		at (\cw+\sw,-4*\rh) {complaint/question};
		
		\node[rectangle, fill=green!4, draw=gray!30, minimum width=\cw, minimum height=\rh,
		font=\scriptsize\sffamily, align=left, inner xsep=6pt, anchor=north west]
		at (0,-5*\rh) {Expansion score $\geq 8$ $\wedge$ system health $H \geq 80$
			\quad {\tiny($p=0.25$)}};
		\node[rectangle, fill=green!4, draw=gray!30, minimum width=\sw, minimum height=\rh,
		font=\scriptsize\sffamily, align=center, anchor=north west]
		at (\cw,-5*\rh) {yes};
		\node[rectangle, fill=green!6, draw=gray!30, minimum width=\tw, minimum height=\rh,
		font=\scriptsize\ttfamily, align=center, anchor=north west]
		at (\cw+\sw,-5*\rh) {feature\_request};
		
		\node[rectangle, fill=red!4, draw=gray!30, minimum width=\cw, minimum height=\rh,
		font=\scriptsize\sffamily, align=left, inner xsep=6pt, anchor=north west]
		at (0,-6*\rh) {Baseline low sentiment $< 0.45$ \quad {\tiny($p=0.15$)}};
		\node[rectangle, fill=red!4, draw=gray!30, minimum width=\sw, minimum height=\rh,
		font=\scriptsize\sffamily, align=center, anchor=north west]
		at (\cw,-6*\rh) {yes};
		\node[rectangle, fill=red!6, draw=gray!30, minimum width=\tw, minimum height=\rh,
		font=\scriptsize\ttfamily, align=center, anchor=north west]
		at (\cw+\sw,-6*\rh) {complaint};
		
		\node[rectangle, fill=gray!8, draw=gray!40, minimum width=\cw, minimum height=\rh,
		font=\scriptsize\itshape\sffamily, align=left, inner xsep=6pt, anchor=north west]
		at (0,-7*\rh) {No condition met};
		\node[rectangle, fill=gray!8, draw=gray!40,
		minimum width=\sw+\tw, minimum height=\rh,
		font=\scriptsize\bfseries\sffamily, text=gray!60, align=center, anchor=north west]
		at (\cw,-7*\rh) {$\varnothing$ \quad Silence is correct, no email fires};
		
		\draw[-{Stealth[length=5pt]}, gray!35, thick]
		(-0.55, -0.05) -- (-0.55, -6.6*\rh)
		node[below, font=\tiny\itshape, text=gray!50] {priority};
		\node[font=\tiny\itshape, text=gray!45, rotate=90, anchor=south]
		at (-0.7, -3.3*\rh) {evaluated in order $\longrightarrow$};
		
	\end{tikzpicture}
	\caption{Signal-driven email decision logic implementing
		Eq.~\ref{eq:email-signals}. Conditions are evaluated top-to-bottom
		in priority order for each customer source $c$ at each simulation
		tick; the first matching condition fires. The terminal $\varnothing$
		row is not a generation gap, it is verifiable ground truth that no
		simulation state warranted contact.}
	\label{fig:email-signals-table}
\end{figure}
	
	The $\varnothing$ case is critical: when no signal condition is met, \emph{no email fires}. This makes absence of customer contact a verifiable ground truth fact, not a gap in generation.
	
	Dropped emails ($\sim$15\%) are still modelled. An \texttt{email\_dropped} SimEvent is emitted with the email artifact ID and a \texttt{no\_action\_taken} reason, creating verifiable communication gaps.
	
	\paragraph{Proactive sales outreach.} After agenda items complete, \texttt{\_fire\_sales\_outreach()} generates daily proactive customer emails from sales team members to their highest-priority open SF opportunities, with cooldown tracking. Combined with \texttt{generate\_customer\_replies()}, which probabilistically generates customer replies that can advance SF deal stages via LLM-assessed \texttt{crm\_stage}, this creates a multi-turn sales conversation cycle with verifiable ground truth stage progression.
	
\subsection{CRM as Organizational Physics}
\label{sec:crm}

The \texttt{CRMSystem} is a full cross-system state machine that extends the
physics-cognition boundary to the customer boundary. A single incident can
trigger a cascade spanning six subsystem boundaries, each emitting a distinct
SimEvent to the ground truth bus (Figure~\ref{fig:crm-cascade}).

On incident resolution, linked Zendesk tickets are closed with postmortem
references. On employee departure, SF accounts and open opportunities are
flagged for reassignment.
	
	\begin{figure}[!t]
		\centering
		\begin{tikzpicture}[
			node distance=0.3cm,
			block/.style={
				rectangle, rounded corners=4pt,
				minimum width=7.2cm, minimum height=0.65cm,
				font=\small\sffamily, align=center,
				inner xsep=8pt, inner ysep=4pt
			},
			simtag/.style={
				font=\tiny\ttfamily, text=gray!55, anchor=west
			},
			arrow/.style={-{Stealth[length=5pt]}, thick, gray!50},
			label/.style={font=\scriptsize\itshape, text=gray!55},
			]

			\node[block, fill=red!8, draw=red!45, text=red!70!black] (trigger) {
				\textbf{System Health Drop} \quad $H\!:\; 85 \;\to\; 42$
			};
			\node[simtag, right=0.25cm of trigger.north east, anchor=north west]
			at (trigger.north east) {\texttt{incident\_opened}};

			\node[block, fill=teal!7, draw=teal!40, below=of trigger] (s1) {
				Datadog alert fires \quad {\small(15-min metric spike recorded)}
			};
			\node[simtag] at (s1.north east) {\texttt{datadog\_alert}};

			\node[block, fill=violet!7, draw=violet!40, below=of s1] (s2) {
				Zendesk tickets escalated to \textbf{Urgent}
				\quad {\small(all open tickets for affected orgs)}
			};
			\node[simtag] at (s2.north east) {\texttt{zd\_tickets\_escalated}};

			\node[block, fill=violet!7, draw=violet!40, below=of s2] (s3) {
				Salesforce opportunities flagged \textbf{at-risk}
				\quad {\small($H < 60$ threshold)}
			};
			\node[simtag] at (s3.north east) {\texttt{sf\_deals\_risk\_flagged}};

			\node[block, fill=teal!5, draw=teal!30, below=of s3] (s4) {
				Daily planning context modified
				\quad {\small\texttt{crm.planner\_context()}}
			};
			\node[simtag] at (s4.north east) {\texttt{day\_plan\_generated}};

			\node[block, fill=teal!5, draw=teal!30, below=of s4] (s5) {
				Voice cards gain CRM pressure
				\quad {\small\texttt{crm\_pressure\_hint()}}
			};
			\node[simtag] at (s5.north east) {\texttt{(prompt-level)}};

			\node[block, fill=orange!8, draw=orange!45, below=of s5] (s6) {
				Signal-driven customer email fires
				\quad {\small(complaint from affected org)}
			};
			\node[simtag] at (s6.north east) {\texttt{inbound\_external\_email}};
			
			\foreach \a/\b in {trigger/s1, s1/s2, s2/s3, s3/s4, s4/s5, s5/s6} {
				\draw[arrow] (\a.south) -- (\b.north);
			}
			
			\coordinate (legend-ref) at ([yshift=-0.6cm]s6.south);
			
			\node[font=\scriptsize\sffamily, text=red!60, anchor=west] 
			at ([xshift=-3.6cm]legend-ref) {\rule{8pt}{6pt}\;Incident trigger};
			
			\node[font=\scriptsize\sffamily, text=teal!70, anchor=west] 
			at ([xshift=-1.1cm]legend-ref) {\rule{8pt}{6pt}\;Engineering};
			
			\node[font=\scriptsize\sffamily, text=violet!70, anchor=west] 
			at ([xshift=1.4cm]legend-ref) {\rule{8pt}{6pt}\;CRM};
			
			\node[font=\scriptsize\sffamily, text=orange!70, anchor=west] 
			at ([xshift=3.7cm]legend-ref) {\rule{8pt}{6pt}\;Email};
			
		\end{tikzpicture}
		\caption{CRM cascade triggered by a single incident crossing six subsystem
			boundaries. Every step emits a SimEvent to the ground truth bus; the cascade
			terminates in a signal-driven customer email whose firing is a verifiable
			consequence of the incident root cause and the affected customer's
			\texttt{depends\_on\_components}. Absence of an email (when no customer
			depends on the affected component) is equally verifiable.}
		\label{fig:crm-cascade}
	\end{figure}
	
	Affected customer orgs are determined by \texttt{\_orgs\_affected\_by\_incident()}, which cross-references each customer's \texttt{depends\_on\_components} (seeded at genesis from the tech stack) against the incident's root cause string. This produces targeted rather than blanket escalation.
	
	\subsection{The Knowledge Recovery Arc}
	\label{sec:knowledge-recovery}
	
	The knowledge recovery arc is a complete organizational knowledge degradation and recovery simulation:
	
	\paragraph{Genesis seeding.} \texttt{seed\_knowledge\_gaps()} creates \texttt{DomainRegistry} entries for each departed employee's knowledge domains, with documentation coverage percentages, system tags for fuzzy matching, and former-owner attribution.
	
	\paragraph{Incremental recovery.} Every Confluence page, PR, and incident resolution that touches an orphaned domain incrementally bumps \texttt{documentation\_coverage}:
	
	\begin{equation}
		\text{coverage}_{d,t+1} = \min\!\left(1.0,\; \text{coverage}_{d,t} + \Delta_{\text{write}}\right)
		\label{eq:coverage-update}
	\end{equation}
	
	where $\Delta_{\text{write}}$ is the coverage delta per write type: $0.10$ for Confluence pages, $0.12$ for design discussion documentation, $0.05$ for async thread documentation of unresolved topics. Matching uses system tags so variant spellings resolve (e.g., ``titan'' matches ``TitanDB'').
	
	\paragraph{Ownership promotion.} Authors are promoted to \texttt{primary\_owner} through three distinct pathways:
	
	\begin{equation}
		\text{promote}(a, d) =
		\begin{cases}
			\text{true} & \text{pathway} = \texttt{confluence} \wedge \text{coverage}_d \geq \theta_{\text{cov}} \\
			\text{true} & \text{pathway} = \texttt{pr} \wedge \text{touches}(a, d) \geq \theta_{\text{pr}} \\
			\text{true} & \text{pathway} = \texttt{incident} \wedge \text{owner}_d = \varnothing \\
			\text{false} & \text{otherwise}
		\end{cases}
		\label{eq:promotion}
	\end{equation}
	
	where $\theta_{\text{cov}} = 0.70$ and $\theta_{\text{pr}} = 3$.
	
	\noindent A deliberate design choice prevents ownership churn: promotion
	fires only when \texttt{primary\_owner} is \texttt{None} (the domain is
	orphaned). An active owner is never displaced by a more prolific
	contributor. This means the knowledge recovery arc is strictly a
	\emph{recovery} process, once ownership is established, it is stable
	unless the new owner themselves departs and triggers a fresh departure
	cascade (Section~\ref{sec:lifecycle}).
	
	\paragraph{Domain claiming on hire.} New hires automatically claim orphaned domains matching their expertise via \texttt{\_claim\_domains\_on\_hire()}.
	
This produces verifiable longitudinal narratives
(Figure~\ref{fig:knowledge-recovery-arc}): e.g., ``Domain X was 20\%
documented when Bill left $\to$ reached 70\% via 3 Confluence pages
$\to$ Priya claimed ownership on Day~15,'' entirely derivable from
the SimEvent log.
	
\begin{figure}[H]
	\centering
	\begin{tikzpicture}[
		x=2.8cm, y=1.0cm,
		event/.style={
			rectangle, rounded corners=4pt, draw=gray!50, fill=white,
			minimum width=2.3cm, minimum height=0.9cm,
			font=\scriptsize\sffamily, align=center, inner sep=4pt
		},
		coverage/.style={
			font=\scriptsize\bfseries\sffamily, text=gray!70
		},
		simtype/.style={
			font=\tiny\ttfamily, text=gray!55
		},
		connector/.style={-{Stealth[length=4pt]}, thick, gray!40},
		daylabel/.style={font=\scriptsize\sffamily, text=gray!60, anchor=north},
		]
	
		\node[event, fill=red!7, draw=red!40, minimum width=2.8cm] (n0) at (0, 0) {
			\textbf{Jordan departs}\\[1pt]
			\simtype{employee\_departed}
		};
		
		\node[event, fill=orange!8, draw=orange!40] (n1) at (1, 0) {
			\textbf{Confluence page}\\touches domain\\[1pt]
			\simtype{confluence\_created}
		};
		
		\node[event, fill=orange!10, draw=orange!50] (n2) at (2, 0) {
			\textbf{Incident surfaces}\\knowledge gap\\[1pt]
			\simtype{knowledge\_gap\_detected}
		};
		
		\node[event, fill=blue!6, draw=blue!35] (n3) at (3, 0) {
			\textbf{PR touches}\\domain\\[1pt]
			\simtype{pr\_reviewed}
		};
		
		\node[event, fill=teal!7, draw=teal!40] (n4) at (4, 0) {
			\textbf{New hire}\\arrives\\[1pt]
			\simtype{employee\_hired}
		};
		
		\node[event, fill=green!7, draw=green!40] (n5) at (5, 0) {
			\textbf{Ownership}\\promoted\\[1pt]
			\simtype{domain\_ownership\_claimed}
		};
		
		\draw[connector] (n0.east) -- (n1.west);
		\draw[connector] (n1.east) -- (n2.west);
		\draw[connector] (n2.east) -- (n3.west);
		\draw[connector] (n3.east) -- (n4.west);
		\draw[connector] (n4.east) -- (n5.west);
		
		\node[daylabel] at (0,  -0.65) {Day 0};
		\node[daylabel] at (1,  -0.65) {Day 3};
		\node[daylabel] at (2,  -0.65) {Day 8};
		\node[daylabel] at (3,  -0.65) {Day 11};
		\node[daylabel] at (4,  -0.65) {Day 15};
		\node[daylabel] at (5,  -0.65) {Day 18};
	
		\node[coverage, text=red!60]   at (0,  0.82) {20\% covered};
		\node[coverage, text=orange!70] at (1, 0.82) {$\to$ 30\%};
		\node[coverage, text=orange!80] at (2, 0.82) {gap logged};
		\node[coverage, text=blue!60]  at (3,  0.82) {$\to$ 45\%};
		\node[coverage] at (4,  0.82)  {domain claimed};
		\node[coverage, text=green!60] at (5,  0.82) {$\to$ 70\%\ $\checkmark$};
		
		\draw[-{Stealth[length=5pt]}, gray!30, thick]
		(-0.5, -1.1) -- (5.6, -1.1);
		
		\node[font=\scriptsize, text=gray!50] at (2.55, -1.35) {simulation time};
		
	\end{tikzpicture}
	\caption{Knowledge recovery arc for a single orphaned domain. Each node represents
		a SimEvent; coverage percentages are derived from the DomainRegistry state at
		the moment of emission. The full arc (from departure through incremental recovery to ownership
		promotion) is entirely derivable from the SimEvent log without post-hoc
		reconstruction.}
	\label{fig:knowledge-recovery-arc}
\end{figure}
	
	\subsection{Multi-Pathway Knowledge Gap Detection}
	\label{sec:gap-detection}
	
	Knowledge gaps are detected through three independent pathways that produce unified \texttt{knowledge\_gap\_detected} events:
	
	\begin{enumerate}[leftmargin=1.5em]
		\item \textbf{Departure-based embedding similarity:} when incident text is semantically similar to a departed employee's expertise vectors (threshold $\geq 0.65$), cross-referenced against the DomainRegistry for live coverage.
		\item \textbf{PR reviewer audit:} reviewers produce structured metadata assessing whether the PR author demonstrates domain competence: \texttt{author\_domain\_fit} (low/medium/high), \texttt{gap\_classification} (none/possible/likely), \texttt{topics\_beyond\_author\_expertise}.
		\item \textbf{Confluence author self-audit:} design doc authors self-assess using the same schema, comparing every topic in their doc against their expertise list.
	\end{enumerate}
	
	All three pathways produce events with identical schema, enabling unified downstream handling.
	
	\subsection{Async Thread Classification and Documentation Spawning}
	\label{sec:async-classification}
	
	Q\&A threads are classified via a fast LLM call as $o \in \{\texttt{resolved}, \texttt{uncertain}, \texttt{unresolved}, \texttt{escalated}\}$. Probability-gated Confluence pages then spawn:
	
	\begin{equation}
		P(\text{spawn} \mid o) =
		\begin{cases}
			0.30 & o = \texttt{escalated} \\
			0.20 & o = \texttt{resolved} \\
			0.10 & o = \texttt{uncertain} \\
			0.05 & o = \texttt{unresolved}
		\end{cases}
		\label{eq:async-doc-prob}
	\end{equation}
	
	Spawned pages update the DomainRegistry, closing the knowledge recovery loop. The most prolific responder in the thread is selected as author.
	
	\subsection{Organizational Lifecycle}
	\label{sec:lifecycle}
	
\subsubsection{Departure Cascade}

When an engineer departs, six steps fire in strict order
(Figure~\ref{fig:departure-cascade}). The ordering is not arbitrary: incident
handoff (Step~1) runs Dijkstra routing while the departing node is still
present in the graph, before any topology changes. Graph recompute (Step~4)
applies proportional stress to engineers absorbing the departed node's bridging
load: $\Delta S_i = \min(20, (g_i^{\text{after}} - g_i^{\text{before}}) \cdot 40)$.
Steps~5--6 seed the knowledge recovery arc
(Section~\ref{sec:knowledge-recovery}).
	
	\begin{figure}[!ht]
		\centering
		\begin{tikzpicture}[
			x=2.6cm, y=1.2cm,
			cascadebox/.style={
				rectangle, rounded corners=3pt, draw=gray!45, fill=white,
				minimum width=2.2cm, minimum height=0.65cm,
				font=\scriptsize\sffamily, align=center, inner sep=3pt
			},
			note/.style={font=\tiny\itshape, text=gray!55, align=left},
			arrow/.style={-{Stealth[length=4pt]}, thick},
			timing/.style={font=\tiny\sffamily, text=gray!60},
			]
						
			\node[cascadebox, fill=red!7, draw=red!40]   (s1) at (0, 0)
			{\textbf{1. Incident handoff}\\Dijkstra reroute};
			\node[note, right=0.3cm of s1] {\emph{Node still in graph}\\escalation path recomputed first};
			
			\node[cascadebox, fill=blue!6, draw=blue!35]  (s2) at (0,-1)
			{\textbf{2. JIRA reassignment}\\orphaned tickets $\to$ lead};
			\node[note, right=0.3cm of s2] {In Progress + no linked PR\\$\to$ reset to To Do};
			
			\node[cascadebox, fill=violet!7, draw=violet!35] (s3) at (0,-2)
			{\textbf{3. CRM ownership lapse}\\SF accounts flagged};
			\node[note, right=0.3cm of s3] {\texttt{sf\_ownership\_lapsed}\\emitted per affected opp};
			
			\node[cascadebox, fill=teal!6, draw=teal!35]  (s4) at (0,-3)
			{\textbf{4. Graph recompute}\\centrality recalculated};
			\node[note, right=0.3cm of s4] {$\Delta S_i = \min(20,\;\Delta g_i \cdot 40)$\\stress bleeds to absorbing nodes};
			
			\node[cascadebox, fill=orange!7, draw=orange!35] (s5) at (0,-4)
			{\textbf{5. Domain orphaning}\\DomainRegistry nulled};
			\node[note, right=0.3cm of s5] {\texttt{primary\_owner} $\to \varnothing$\\new gap entries created};
			
			\node[cascadebox, fill=orange!10, draw=orange!45] (s6) at (0,-5)
			{\textbf{6. Knowledge gap propagation}\\gaps logged to SimEvent bus};
			\node[note, right=0.3cm of s6] {\texttt{knowledge\_gap\_detected}\\per untracked domain};
						
			\foreach \a/\b in {s1/s2, s2/s3, s3/s4, s4/s5, s5/s6}{
				\draw[arrow, gray!50] (\a.south) -- (\b.north);
			}
						
			\draw[dashed, red!40, thick]
			($(s1.north west)+(-0.12, 0.15)$) rectangle ($(s1.south east)+(0.12,-0.12)$);
			\node[font=\tiny\itshape, text=red!55, anchor=west]
			at ($(s1.north east)+(2.0, 0.0)$)
			{\(\leftarrow\) must fire \emph{before} node removal};
			
		\end{tikzpicture}
		\caption{Ordered departure cascade. Steps execute sequentially; the critical
			ordering constraint is that incident handoff (Step~1) uses Dijkstra routing
			\emph{while the departing node is still present in the graph}, before any
			topology changes. Steps~4--6 propagate the structural consequences (stress redistribution,
			domain orphaning, and knowledge gap registration) which seed the knowledge
			recovery arc (Section~\ref{sec:knowledge-recovery}).}
		\label{fig:departure-cascade}
	\end{figure}
	
	\subsubsection{Automated Backfill}
	
	\texttt{\_schedule\_backfill()} queues a replacement hire after a configurable lag (default 14 days). \texttt{\_generate\_backfill\_persona()} asks the LLM to generate a full persona (name, expertise, style, social role, typing quirks) constrained to not collide with existing names. On arrival, the new hire enters the graph with cold-start edges ($2\times\omega_{\min}$ intra-department, $\omega_{\min}$ cross-department), claims orphaned domains matching their expertise, and naturally attracts 1-on-1s and mentoring sessions from the planner.
	
	\subsection{Causal Timestamp Consistency}
	\label{sec:simclock}
	
	An actor-local clock in which every employee maintains an independent time cursor eliminates causal timestamp violations. Two core primitives:
	
	\begin{itemize}[leftmargin=1.5em]
		\item \textbf{\texttt{advance\_actor($i$, $\Delta$)}} --- models parallel work. Advances only actor $i$'s cursor.
		\item \textbf{\texttt{sync\_and\_tick($A$, $\Delta$)}} --- models causal chains. Synchronises all participants to $\max_{i \in A}(\text{cursor}_i)$, then advances. Guarantees no response artifact receives a timestamp earlier than its trigger.
	\end{itemize}

\section{Corpus Properties}

The architectural mechanisms described in Section~\ref{sec:architecture}
produce corpora with four properties that existing synthetic datasets
lack simultaneously: cross-artifact causal traceability, temporal structure,
verified absence as ground truth, and configurable organizational complexity.

\subsection{Cross-Artifact Causal Traceability}
\label{sec:causal-traceability}

Every artifact produced by OrgForge carries the \texttt{incident\_id} of the
SimEvent that caused it, enabling the \texttt{CausalChainHandler} to
reconstruct full causal chains from the ground truth bus without post-hoc
inference. A single infrastructure incident produces a traceable chain spanning
both the engineering resolution path and the simultaneous CRM and customer
impact path (Figure~\ref{fig:incident-trace}).

\begin{figure}[!ht]
	\centering
	\begin{tikzpicture}[
		node distance=0.18cm and 0.1cm,
		artifact/.style={
			rectangle, rounded corners=4pt,
			minimum width=2.35cm, minimum height=0.72cm,
			font=\scriptsize\sffamily, align=center,
			draw, inner xsep=3pt, inner ysep=4pt
		},
		arrow/.style={-{Stealth[length=4pt]}, thick},
		simlabel/.style={font=\tiny\ttfamily, text=gray!50},
		timelabel/.style={font=\tiny\sffamily, text=gray!55, anchor=north},
		]

		\node[artifact, fill=red!8,    draw=red!45,    text=red!70!black]
		(trigger) at (0, 0)
		{\textbf{Datadog}\\metric spike};
		
		\node[artifact, fill=red!10,   draw=red!55,    text=red!75!black,
		right=0.45cm of trigger]
		(inc) {\textbf{Incident}\\opened};
		
		\node[artifact, fill=teal!7,   draw=teal!40,
		right=0.45cm of inc]
		(slack) {\textbf{Slack}\\alert thread};
		
		\node[artifact, fill=blue!7,   draw=blue!35,
		right=0.45cm of slack]
		(jira) {\textbf{JIRA}\\ticket created};
		
		\node[artifact, fill=blue!10,  draw=blue!45,
		right=0.45cm of jira]
		(pr) {\textbf{GitHub PR}\\fix merged};
		
		\node[artifact, fill=teal!9,   draw=teal!50,
		right=0.45cm of pr]
		(conf) {\textbf{Confluence}\\postmortem};

		\node[artifact, fill=violet!7, draw=violet!40,
		below=1.05cm of inc]
		(zd) {\textbf{Zendesk}\\ticket $\to$ Urgent};
		
		\node[artifact, fill=violet!9, draw=violet!50,
		right=0.45cm of zd]
		(sf) {\textbf{Salesforce}\\opp risk-flagged};
		
		\node[artifact, fill=orange!8, draw=orange!40,
		right=0.45cm of sf]
		(email) {\textbf{Inbound email}\\customer complaint};
		
		\node[artifact, fill=orange!10,draw=orange!50,
		right=0.45cm of email]
		(inv) {\textbf{Invoice}\\SLA credit applied};
		
		\node[artifact, fill=gray!12,  draw=gray!45,
		right=0.45cm of inv]
		(nps) {\textbf{NPS survey}\\score degraded};

		\foreach \a/\b in {trigger/inc, inc/slack, slack/jira, jira/pr, pr/conf}{
			\draw[arrow, gray!50] (\a.east) -- (\b.west);
		}

		\draw[arrow, violet!50]
		(inc.south) -- ++(0,-0.38cm) -| (zd.north);

		\foreach \a/\b in {zd/sf, sf/email, email/inv, inv/nps}{
			\draw[arrow, gray!50] (\a.east) -- (\b.west);
		}

		\node[font=\tiny\bfseries\sffamily, text=teal!70!black, rotate=90,
		anchor=center]
		at (-1.4cm, 0) {Engineering};
		
		\node[font=\tiny\bfseries\sffamily, text=violet!65!black, rotate=90,
		anchor=center]
		at (-1.4cm, -2.08cm) {CRM};

		\draw[dashed, gray!30, thin]
		(-0.7cm, -0.62cm) -- (15.2cm, -0.62cm);

		\node[font=\tiny\itshape\sffamily, text=gray!50, anchor=west]
		at (pr.south east |- conf.south)
		{};
		\draw[decorate,
		decoration={brace, amplitude=4pt, mirror},
		gray!35]
		([yshift=-0.05cm]trigger.south west) --
		([yshift=-0.05cm]conf.south east)
		node[midway, below=0.18cm,
		font=\tiny\itshape\sffamily, text=gray!50]
		{same \texttt{incident\_id} stamped on every artifact};
		
	\end{tikzpicture}
	\caption{Cross-artifact causal chain produced by a single infrastructure
		incident. The top row shows the engineering resolution path; the bottom row
		shows the simultaneous CRM and customer impact path. Every artifact carries
		the same \texttt{incident\_id} in its SimEvent payload, enabling the
		\texttt{CausalChainHandler} to reconstruct the full chain from the ground truth
		bus. The branch from incident to Zendesk fires via
		\texttt{\_orgs\_affected\_by\_incident()}, which cross-references each
		customer's \texttt{depends\_on\_components} against the incident root cause;
		customers whose stack does not include the affected component produce no
		bottom-row artifacts, a verifiable absence (Section~\ref{sec:email}).}
	\label{fig:incident-trace}
\end{figure}
	
	\subsection{Temporal Structure}
	
	The $N$-day simulation produces facts with known temporal validity windows. System health degrades on incident opening and recovers on resolution. Domain registry coverage increases incrementally through documentation. Edge weights evolve continuously. Deal stages advance through customer reply cycles. This temporal structure enables questions requiring temporal reasoning rather than simple lookup.
	
	\subsection{Verified Absence as Ground Truth}
	
	The signal-driven email model (Eq.~\ref{eq:email-signals}) means silence is verifiable. If a customer with \texttt{depends\_on\_components: ["Kafka", "PostgreSQL"]} did not email during an incident affecting only Redis, that absence is ground truth, not a gap in generation. Similarly, \texttt{email\_dropped} events create verifiable communication failures.
		
	\subsection{Configurable Organizational Complexity}
	
	\texttt{config.yaml} specifies company name, industry, organizational structure, persona definitions, incident triggers, CRM configuration, and lifecycle events. Teams of 5 to 50+ can be simulated.

\subsection{Corpus Reproducibility}
\label{sec:reproducibility}

OrgForge produces three layers with distinct reproducibility properties.

\paragraph{Tier 1: The structural event skeleton is deterministic.}
Given identical \texttt{config.yaml} and random seed, the engine produces an
identical sequence of lifecycle events (departures, hires), incident timing
(via the seeded probability path), sprint cadence, on-call rotation, CRM
cascades, signal-driven email decisions (Equation~\ref{eq:email-signals}),
and all GraphDynamics computations (Equations~\ref{eq:stress-propagation}--\ref{eq:crm-edge-sync}).
These events form the ground truth bus and are reproducible across runs.

\paragraph{Tier 2: Fine-grained daily activity is planner-influenced.}
The specific agenda items dispatched per engineer depend on LLM planner
outputs ($P$): the exact distribution of \texttt{async\_question},
\texttt{design\_discussion}, and \texttt{1on1} SimEvents may vary between
runs. Ticket titles generated during sprint planning influence
embedding-based assignment scores (Equation~\ref{eq:assignment-score}).
However, all activities are validated by $V$ before execution and
constrained to the engine-owned roster, ticket ownership, and capacity
model (Equation~\ref{eq:capacity}). The structural invariants: who is
on-call, which domains are orphaned, which customers are affected, are
identical across runs.

\paragraph{Tier 3: Prose artifacts are not reproducible.}
Slack messages, Confluence pages, emails, and meeting transcripts are
generated by the LLM from validated SimEvent context. Two runs from the
same seed produce structurally equivalent corpora: identical Tier~1
facts, causal chains, and actor assignments, with different Tier~2
activity distributions and different surface prose.

\medskip\noindent
This distinction has a direct implication for evaluation design:
\textbf{evaluation queries must be answerable from Tier~1 SimEvents.}
The Tier~1 log is the answer key; Tier~2 and Tier~3 artifacts are the
retrieval surface. A researcher who re-generates from a published
Tier~1 log receives a corpus valid for all the same evaluation queries.	

	\subsection{Corpus Statistics: 60-Day Reference Run}
\label{sec:corpus-stats}

Tables~\ref{tab:corpus-stats} and~\ref{tab:org-config} report artifact counts, SimEvent
counts, and organizational configuration for the reference 60-day run.

\begin{table}[!t]
	\centering
	\small
	\begin{minipage}[t]{0.42\textwidth}
		\centering
		\begin{tabular}{lr}
			\toprule
			\textbf{Artifact type} & \textbf{Count} \\
			\midrule
			Slack threads       & 3{,}158 \\
			Email (all)         & 552     \\
			Confluence pages    & 474     \\
			JIRA tickets        & 328     \\
			Zoom transcripts    & 193     \\
			Pull requests       & 53      \\
			Datadog alerts      & 11      \\
			Zendesk tickets     & 8       \\
			Invoices            & 8       \\
			Salesforce opps     & 2       \\
			NPS surveys         & 1       \\
			\midrule
			\textbf{Total}      & \textbf{4{,}788} \\
			\bottomrule
		\end{tabular}
	\end{minipage}
	\hfill
	\begin{minipage}[t]{0.54\textwidth}
		\centering
		\begin{tabular}{lr}
			\toprule
			\textbf{SimEvent type} & \textbf{Count} \\
			\midrule
			\texttt{datadog\_metric}           & 5{,}760 \\
			\texttt{knowledge\_gap\_detected}  & 2{,}106 \\
			\texttt{deep\_work\_session}       & 2{,}060 \\
			\texttt{async\_question}           & 1{,}465 \\
			\texttt{confluence\_created}       & 468     \\
			\texttt{design\_discussion}        & 443     \\
			\texttt{dept\_plan} (all variants) & 1{,}260 \\
			\texttt{1on1}                      & 368     \\
			\texttt{watercooler\_chat}         & 357     \\
			\texttt{ticket\_progress}          & 356     \\
			\texttt{jira\_ticket\_created}     & 296     \\
			\texttt{inbound\_external\_email}  & 277     \\
			\texttt{mentoring}                 & 244     \\
			All other types (32)               & 1{,}332 \\
			\midrule
			\textbf{Total}                     & \textbf{16{,}792} \\
			\bottomrule
		\end{tabular}
	\end{minipage}
	\caption{Artifact and SimEvent counts for the 60-day reference run. The 32
		lower-frequency SimEvent types include \texttt{incident\_opened} ($n=11$),
		\texttt{employee\_departed} ($n=5$), \texttt{email\_dropped} ($n=6$), and
		\texttt{blocker\_flagged} ($n=1$), among others. The full per-type breakdown
		is available in the published SimEvent log on Hugging Face.}
	\label{tab:corpus-stats}
\end{table}

\begin{table}[!t]
	\centering
	\small
	\begin{tabular}{llr}
		\toprule
		\textbf{Parameter} & \textbf{Detail} & \textbf{Value} \\
		\midrule
		\multicolumn{3}{l}{\textit{Organizational scale}} \\
		\quad Employees          & Active at Day 0                  & 41 \\
		\quad Vendor sources     & Seeded at genesis                & 7  \\
		\quad Customer accounts  & Seeded at genesis                & 8  \\
		\midrule
		\multicolumn{3}{l}{\textit{Lifecycle events}} \\
		\quad Departures (in-simulation)  & Days 1--60                        & 4  \\
		\quad Backdated departure  & Day of departure (relative to Day 0) & $-$639 \\
		\quad \quad \textit{(genesis-seeded knowledge gap)} &
		\textit{Domain orphan age at Day 0} & \textit{639 days} \\
		\quad New hires (in-simulation)   & Days 1--60                        & 4  \\
		\bottomrule
	\end{tabular}
	\caption{Organizational configuration for the 60-day reference run. The backdated
		departure was seeded at genesis rather than emitted as an in-simulation
		\texttt{employee\_departed} SimEvent: its domains enter the DomainRegistry at Day~0
		already orphaned for 639 days ($\approx$21 months), producing a knowledge gap
		of significantly greater depth than any gap created within the run window. This
		is the most information-sparse scenario the knowledge recovery arc can encounter
		and represents a realistic long-term attrition case.}
	\label{tab:org-config}
\end{table}

\subsection{Cross-Document Consistency Evaluation}
\label{sec:consistency-eval}

We evaluate OrgForge artifacts against two LLM-only baselines on two metrics
computed across $N = 10$ incidents (${\approx}20$ artifacts each).

\paragraph{Arms.}
\begin{description}[leftmargin=1.5em, labelwidth=5em, font=\normalfont\bfseries]
	\item[OrgForge]   Full simulation pipeline with the SimEvent ground-truth bus.
	\item[Chained]    All five artifact types generated from the same incident brief
	in a single chained LLM context; each document receives all
	prior documents as context.
	\item[Parallel]   Each artifact type generated independently from the incident
	brief with no cross-document context.
\end{description}

\paragraph{Metrics.}
\textit{Entity agreement} measures grounded precision of tech-component, person-name, and
ticket-ID mentions against the SimEvent actor and causal-chain records (higher is better).
\textit{Prose-SimEvent fidelity} is a weighted composite
$\Delta = 0.35\,s_{\text{ent}} + 0.45\,s_{\text{nli}} + 0.20\,s_{\text{num}}$
measuring alignment between artifact prose and the corresponding SimEvent
ground-truth facts, where $s_{\text{ent}}$ is entity recall, $s_{\text{nli}}$
is NLI-based entailment, and $s_{\text{num}}$ is numeric consistency (higher is better).
Temporal ordering violations are detectable from the SimEvent log directly and
are therefore a property of the simulation architecture rather than an empirical
metric over generated artifacts; we omit them from the comparison table.

\begin{table}[!t]
	\centering
	\small
	\setlength{\tabcolsep}{8pt}
	\begin{tabular}{lrrr}
		\toprule
		\textbf{Metric}
		& \textbf{OrgForge}
		& \textbf{Chained}
		& \textbf{Parallel} \\
		\midrule
		Entity agreement $\uparrow$
		& $\mathbf{0.9920 \pm 0.0030}$ & $0.8498 \pm 0.0663$ & $0.6469 \pm 0.0301$ \\
		Prose-SimEvent fidelity $\uparrow$
		& $\mathbf{0.9962 \pm 0.0038}$ & $0.5396 \pm 0.0625$ & $0.5315 \pm 0.0730$ \\
		\midrule
		Incidents evaluated & 10 & 10 & 10 \\
		Artifacts per incident (mean $\pm$ sd) & $19.5 \pm 0.7$ & $5.0$ & $5.0$ \\
		\bottomrule
	\end{tabular}
	\caption{Cross-document consistency across three arms (mean $\pm$ 95\% CI,
		$t$-distribution, $df=9$). Entity agreement for OrgForge is grounded
		precision against the SimEvent actor and causal-chain records; for
		baselines, pairwise Jaccard agreement across artifact mentions.
		Prose-SimEvent fidelity measures alignment between artifact prose and the
		ground-truth fact set: for OrgForge, against the \texttt{incident\_opened}
		SimEvent; for baselines, against the incident brief used as generation
		input. The overlapping CIs for Chained and Parallel on fidelity
		($[0.477, 0.602]$ vs.\ $[0.459, 0.604]$) confirm that chaining produces
		no statistically distinguishable improvement in factual correctness.
		Bold denotes best per row.}
	\label{tab:consistency-eval}
\end{table}

\paragraph{The consistent hallucination problem.}
Chaining prior artifacts into the prompt improves cross-document entity
agreement relative to the parallel baseline ($0.8498$ vs.\ $0.6469$, $\Delta =
+0.20$), confirming that context propagates entity mentions reliably. However,
chaining produces no corresponding improvement in factual fidelity against
ground truth ($0.5396$ vs.\ $0.5315$, $\Delta = 0.008$). This dissociation
isolates the \emph{consistent hallucination} failure mode: if an artifact early
in the chain fabricates a root cause or entity, chaining propagates that
fabrication faithfully into subsequent artifacts, producing a corpus that is
internally coherent but factually wrong. The 0.46 absolute gap between
\textsc{OrgForge} (0.9962) and both baselines (${\approx}0.54$) on
Prose-SimEvent fidelity validates structured fact injection through the
physics-cognition boundary as the operative mechanism. Internal consistency is
a consequence of coherent generation; factual fidelity requires an external
ground-truth enforcer.

\paragraph{Asymmetry in evaluation difficulty.}
OrgForge incidents yielded a mean of 19.5 artifacts ($\sigma = 0.7$) spanning
five document types (JIRA tickets, Slack threads, pull requests, Confluence
pages, and postmortems), compared to exactly 5 artifacts per baseline incident.
Entity agreement is computed over all $\binom{k}{2}$ artifact pairs per
incident, yielding approximately 1,900 pairwise comparisons for OrgForge versus
100 for each baseline across the full evaluation set. Achieving $0.9920$
entity agreement across a 20-artifact heterogeneous bundle --- in which JIRA
comments, Slack threads, and PRs naturally mention disjoint entity subsets ---
is a materially harder task than scoring agreement across 5 artifacts of the
same type. This asymmetry strengthens rather than weakens the claim: the
comparison is not apples-to-apples, and the directional disadvantage falls on
\textsc{OrgForge}.

Table~\ref{tab:consistency-divergence} breaks the Prose-SimEvent divergence
composite into its three components, restricted to the OrgForge arm where
ground-truth SimEvent facts are available.

\begin{table}[H]
	\centering
	\small
	\setlength{\tabcolsep}{8pt}
	\begin{tabular}{lrrrr}
		\toprule
		\textbf{Artifact type}
		& $s_{\text{ent}}$ $\uparrow$
		& $s_{\text{nli}}$ $\uparrow$
		& $s_{\text{num}}$ $\uparrow$
		& $\Delta$ (composite) $\uparrow$ \\
		\midrule
		JIRA ticket      & $0.950 \pm 0.113$ & $1.000$ & $1.000$ & $0.983 \pm 0.040$ \\
		Pull request     & $1.000$ & $1.000$ & $1.000$ & $1.000$ \\
		Confluence page  & $1.000$ & $0.975 \pm 0.025$ & $1.000$ & $0.989 \pm 0.011$ \\
		\midrule
		\textit{Mean}    & $0.983 \pm 0.034$ & $0.992 \pm 0.009$ & $1.000$ & $0.990 \pm 0.012$ \\
		\bottomrule
	\end{tabular}
	\caption{Prose-SimEvent divergence component scores for OrgForge artifacts,
		averaged over 10 incidents. $s_{\text{ent}}$: entity recall against SimEvent
		actor and identifier fields. $s_{\text{nli}}$: NLI entailment score using
		DeBERTa-v3-base-mnli-fever-anli~\citep{laurer2024deberta-nli,he2023debertav3}
		with fact templates. $s_{\text{num}}$:
		numeric consistency within a 15\% relative tolerance.
		Composite $\Delta = 0.35\,s_{\text{ent}} + 0.45\,s_{\text{nli}} + 0.20\,s_{\text{num}}$.
		Intervals are 95\% CIs (Student's $t$, $n=10$); cells without intervals had zero variance across all incidents.}
	\label{tab:consistency-divergence}
\end{table}

\section{Enabled Evaluation Surfaces}
\label{sec:eval-surfaces}

Table~\ref{tab:eval-surfaces} maps each evaluation surface to the architectural
subsystems that are necessary preconditions for it and provides a representative
query. All surfaces additionally require the SimEvent bus as the answer-key layer.
Formal benchmarks, scoring equations, and leaderboards are deferred to the forthcoming companion evaluation paper.

\begin{table}[!t]
	\centering
	\setlength{\tabcolsep}{4pt}
	\renewcommand{\arraystretch}{1.45}
	\resizebox{\textwidth}{!}{%
		\begin{tabular}{lp{5.2cm}ccccccc}
			\toprule
			\textbf{Evaluation Surface}
			& \textbf{Representative Query}
			& \rotatebox{65}{\parbox{2.4cm}{\centering\textcolor{teal!70!black}{SimEvent\\bus}}}
			& \rotatebox{65}{\parbox{2.4cm}{\centering\textcolor{violet!70!black}{CRM\\cascade}}}
			& \rotatebox{65}{\parbox{2.4cm}{\centering\textcolor{orange!70!black}{Email\\signals}}}
			& \rotatebox{65}{\parbox{2.4cm}{\centering\textcolor{blue!70!black}{Knowledge\\recovery}}}
			& \rotatebox{65}{\parbox{2.4cm}{\centering\textcolor{red!60!black}{Departure\\cascade}}}
			& \rotatebox{65}{\parbox{2.4cm}{\centering\textcolor{teal!60!black}{Graph\\dynamics}}}
			& \rotatebox{65}{\parbox{2.4cm}{\centering\textcolor{gray!70}{Voice\\cards}}} \\
			\midrule
			Longitudinal narrative reconstruction
			& \emph{``Trace the ownership history of the TitanDB domain from Day~0 to Day~18.''}
			& \checkmark & & & \checkmark & \checkmark & \checkmark & \\
			
			Cross-system causal cascade
			& \emph{``Why does the Acme Corp invoice include a \$200 SLA credit?''}
			& \checkmark & \checkmark & \checkmark & & & & \\
			
			Verified absence reasoning
			& \emph{``Which customers were affected by the Day~8 incident but never contacted support?''}
			& \checkmark & & \checkmark & & & & \\
			
			Actor-scoped epistemic reasoning
			& \emph{``Could Morgan have known the Acme deal was at risk on Day~14 at 14:00?''}
			& \checkmark & \checkmark & & & & & \checkmark \\
			
			Multi-department strategic synthesis
			& \emph{``Synthesize the Acme relationship from Engineering, Sales, and Support perspectives.''}
			& \checkmark & \checkmark & & & & & \checkmark \\
			
			Process compliance auditing
			& \emph{``Were all PRs in sprint~3 reviewed before merge?''}
			& \checkmark & & & & \checkmark & & \\
			
			Counterfactual organizational reasoning
			& \emph{``Would the escalation chain have been shorter if Jordan had documented auth-service?''}
			& \checkmark & & & \checkmark & \checkmark & \checkmark & \\
			\bottomrule
		\end{tabular}%
	}
	\caption{Evaluation surfaces enabled by OrgForge corpora mapped to their architectural
		preconditions. A \checkmark\ indicates the surface depends on the correctness and
		completeness of that subsystem. Benchmark construction, scoring equations, and
		multi-run statistical analysis are deferred to the forthcoming companion evaluation paper.}
	\label{tab:eval-surfaces}
\end{table}

\subsection{Worked Example: Cross-System Causal Cascade}
\label{sec:worked-example}

We trace the representative query \emph{``Why does the Acme Corp invoice
	include a \$200 SLA credit?''} through the SimEvent chain that constitutes
its ground-truth answer and the prose documents a RAG system must retrieve.

\paragraph{Ground-truth answer (SimEvent chain).}
The answer requires backward traversal through six SimEvents, each carrying
the same \texttt{incident\_id} (Figure~\ref{fig:incident-trace}):

\begin{enumerate}[leftmargin=1.5em, itemsep=2pt]
	\item \texttt{incident\_opened} (Day~8, 09:15) --- system health drops
	$85 \to 42$; root cause: \texttt{redis-cluster-split}. Incident
	duration: 62.3\,h.
	
	\item \texttt{zd\_tickets\_escalated} (Day~8, 09:18) ---
	\texttt{\_orgs\_affected\_by\_incident()} cross-references Acme
	Corp's \texttt{depends\_on\_components: ["Redis", "Kafka"]} against
	the root cause; match on Redis triggers Zendesk escalation to
	\textbf{Urgent}.
	
	\item \texttt{sf\_deals\_risk\_flagged} (Day~8, 09:20) --- system health
	$< 60$ threshold flags Acme Corp's open renewal opportunity as
	at-risk.
	
	\item \texttt{inbound\_external\_email} (Day~8, 11:42) --- signal cascade
	(Eq.~\ref{eq:email-signals}, row~1) fires a complaint email from
	Acme Corp's primary contact.
	
	\item \texttt{incident\_resolved} (Day~10, 23:32) --- total downtime
	62.3\,h recorded in SimEvent payload.
	
	\item \texttt{invoice\_generated} (post-simulation) --- SLA terms apply a
	credit of $r_{\text{SLA}} = 2\%$ of monthly recurring revenue per breach
	day, where a breach day is any calendar day the incident spans beyond a
	one-day resolution threshold ($\tau = 1$\,d). Acme Corp's contract ARR is
	\$60{,}000, giving $\text{MRR} = \$5{,}000$. The incident spans
	$d = 3$ calendar days, so $\text{breach\_days} = \max(0,\, 3 - 1) = 2$
	and the credit is:
	\[
	\text{credit} = \text{MRR} \times r_{\text{SLA}} \times \text{breach\_days}
	= \$5{,}000 \times 0.02 \times 2 = \$200.
	\]
	The SLA credit line item in the invoice JSON carries
	\texttt{breach\_days: 2}, \texttt{credit\_rate: 0.02}, and
	\texttt{amount: -200.00}, traceable to \texttt{incident\_id: ENG-\{n\}} via
	the \texttt{InvoiceWriter} in \path{post_sim_artifacts.py}.
\end{enumerate}

\noindent Every link in this chain is a SimEvent with a timestamp, actor
set, and \texttt{incident\_id}; no link requires reading prose.

\paragraph{Retrieval surface (prose documents).}
A RAG system answering this query must retrieve \emph{at least} three of the
following artifacts, each generated by an LLM from the corresponding SimEvent
context:

\begin{itemize}[leftmargin=1.5em, itemsep=2pt]
	\item The Datadog alert Slack message in \texttt{\#system-alerts} (Day~8,
	09:16).
	\item The Zendesk ticket comment thread showing escalation to Urgent.
	\item The inbound complaint email from Acme Corp (\texttt{.eml} file).
	\item The incident postmortem Confluence page (Day~9), which names
	\texttt{redis-cluster-split} as root cause.
	\item The SLA-adjusted invoice JSON (post-simulation derived).
\end{itemize}

\noindent The difficulty is that no single document contains the full
answer. The invoice states the credit amount but not the root cause; the
postmortem states the root cause but not the credit; the email confirms
customer impact but not the SLA calculation. Only by retrieving and
synthesizing across artifact types can the query be answered correctly ---
and the SimEvent chain provides the verifiable ground truth to score
whether the answer is complete.
	
	\section{Future Work}
	
	\paragraph{Non-engineering company simulation.}
	The current execution layer has a structural \texttt{eng/non\_eng} binary. A completion pathway abstraction, configurable review cycles for legal briefs, clinical treatment plans, or logistics shipment approvals, would make OrgForge the first simulation system producing verifiable organizational corpora for any industry.
	
	\paragraph{Configurable incident archetypes.}
	The incident model currently assumes infrastructure failures. Configurable archetypes (regulatory findings, missed filing deadlines, patient safety events) would extend the simulation to regulated industries.
	
	\paragraph{Multi-stakeholder customer contacts.}
	The current CRM model has one contact per customer org. Multi-stakeholder contacts (clinical champion, procurement lead, CISO) with different communication patterns would create richer customer corpora.
	
	\paragraph{Domain packs.}
	Pre-configured \texttt{config.yaml} templates for healthcare, fintech, and legal domains would substantiate the domain-agnosticism claim with demonstrated examples.
	
	\paragraph{Plugin architecture.}
	A formal plugin interface for community-contributed artifact types (PagerDuty, Linear, Looker dashboards) remains on the roadmap.
	
	\paragraph{Formal evaluation benchmarks.}
	The evaluation surfaces described in Section~\ref{sec:eval-surfaces} define the properties; building rigorous benchmarks with scoring equations, multi-run statistical analysis, and leaderboards against these properties is a natural next step for the community.
	
\section{Conclusion}

We have presented OrgForge, a multi-agent simulation framework for generating
synthetic organizational corpora with verifiable ground truth. The central
contribution is an architectural boundary formalized as $M = (S, P, V, E)$
that separates fact control from prose generation, making internal consistency
an architectural guarantee rather than an emergent property.

OrgForge does not simulate documents, it simulates the organizational
processes that produce documents. The knowledge recovery arc produces
verifiable longitudinal narratives of organizational knowledge degradation
and recovery. The signal-driven customer email model makes silence verifiable
ground truth. The CRM state machine extends the physics-cognition boundary
to the customer boundary, producing cross-system causal cascades spanning
six subsystem boundaries. The embedding-based ticket assignment system makes
the simulation domain-agnostic. Non-engineering department simulation produces
heterogeneous completion artifacts with full causal chain parity. 

The combination of these process simulations produces corpora with properties
that no existing synthetic dataset provides: verified absence, longitudinal
organizational narratives, cross-system cascade traceability, and
department-heterogeneous artifacts with full ground truth. These properties
serve a broader surface than evaluation alone. Training pipelines benefit from
the same cross-document consistency guarantee that benchmarks require.
Organizational agents can be tested against a live simulation with verifiable
ground truth before deployment. Security and compliance tooling gains
cross-system data with known labels that real corpora cannot provide without
legal constraint. And the GraphDynamics subsystem and knowledge recovery arc
stand as independent tools for organizational behavior research. OrgForge
is infrastructure for any system that requires organizational ground truth
to be guaranteed rather than assumed.

	\section*{Acknowledgments}
	
	This work was conducted independently without external funding or institutional support.

\end{document}